\documentclass{article}

\usepackage{
    algorithm,
    algpseudocode,
    amsfonts,
    amsmath,
    amssymb,
    booktabs,
    enumitem,
    graphicx,
    hyperref,
    microtype,
    microtype,
    multicol,
    multirow,
    nicefrac,
    xcolor,
    siunitx,
    subcaption,
    tabularx,
    times,
    todonotes,
    url,
    xspace,
    marginnote,
}
\usepackage[
]{caption}
\usepackage[capitalize,nameinlink,noabbrev]{cleveref}
\usepackage[T1]{fontenc}
\usepackage[utf8]{inputenc}
\usepackage[most,breakable]{tcolorbox}
\usepackage{titletoc}
\usepackage{placeins}
\usepackage[table]{xcolor}
\usepackage{etoolbox}

\newtcolorbox{promptbox}[2][]{%
  enhanced,breakable,
  colback=gray!5, colframe=black, fontupper=\scriptsize\ttfamily,
  title={\figurename~\refstepcounter{figure}\thefigure: #2},
  #1
}

\usepackage{iclr2026_conference}
\iclrfinalcopy

\usepackage{wrapfig}

\definecolor{accentA}{HTML}{2563EB}
\definecolor{accentB}{HTML}{7C3AED}
\definecolor{softBg}{HTML}{F8FAFC}
\definecolor{softFrame}{HTML}{E2E8F0}

\renewcommand{\paragraph}[1]{\textbf{#1}\hspace{0.2cm}}

\crefformat{equation}{Eq.~#2\textup{#1}#3}
\crefformat{figure}{Fig.~#2\textup{#1}#3}
\crefformat{table}{Tab.~#2\textup{#1}#3}
\crefformat{appendix}{App.~#2\textup{#1}#3}

\crefformat{section}{\S#2#1#3}
\crefmultiformat{section}{\S\S#2#1#3}{ and~#2#1#3}{, #2#1#3}{, and~#2#1#3}

\tcbset{boxrule=0.6pt, arc=3mm, enhanced, frame style={opacity=0.9}}
\newtcolorbox{profilebox}{
  colback=softBg, colframe=softFrame,
  title=\textbf{User Preference Profile}, coltitle=black,
  fonttitle=\bfseries\small, drop fuzzy shadow,
  left=2pt,right=2pt,top=2pt,bottom=2pt,boxsep=2pt
}
\newtcolorbox{qcardA}[1]{
  title=\textbf{#1}, fonttitle=\bfseries\small,
  colback=white, colframe=accentA!40, colbacktitle=accentA!8, coltitle=black,
  drop fuzzy shadow, left=2pt, right=2pt, top=1pt, bottom=1pt, boxsep=1pt,
  toptitle=1pt, bottomtitle=1pt
}
\newtcolorbox{qcardB}[1]{
  title=\textbf{#1}, fonttitle=\bfseries\small,
  colback=white, colframe=accentB!40, colbacktitle=accentB!8, coltitle=black,
  drop fuzzy shadow, left=2pt, right=2pt, top=1pt, bottom=1pt, boxsep=1pt,
  toptitle=1pt, bottomtitle=1pt
}
\newtcolorbox{metricA}{colback=accentA!8, colframe=accentA!55, arc=2mm, boxrule=0.6pt,
  left=1pt, right=1pt, top=2pt, bottom=2pt, boxsep=1pt}
\newtcolorbox{metricB}{colback=accentB!8, colframe=accentB!55, arc=2mm, boxrule=0.6pt,
  left=1pt, right=1pt, top=2pt, bottom=2pt, boxsep=1pt}

\newlength{\CardH}
\hypersetup{
    colorlinks,
    linkcolor={blue!50!black},
    citecolor={blue!50!black},
    urlcolor={blue!50!black},
}
\newcommand{\name}{BED-LLM\xspace}
\newcommand{\baseline}{Prompt-Only\xspace}

\newcommand{\eig}{\mathrm{EIG}}
\newcommand{\E}{\mathbb{E}}

\newcommand{\design}{x}
\newcommand{\data}{y}

\newcommand{\argmax}{\mathop{\mathrm{arg\,max}}}
\newcommand{\argmin}{\mathop{\mathrm{arg\,min}}}

\newcommand{\eigt}{\eig_\theta}

\newcommand{\history}{h_{t-1}}
\newcommand{\model}{p(\theta;\history)}
\newcommand{\prior}{p(\theta)}
\newcommand{\jointmodel}{p(\theta,\data;\design)}
\newcommand{\marginal}{p(\data;\design)}
\newcommand{\posterior}{p(\theta|\data;\design)}
\newcommand{\likelihood}{p(\data|\theta;\design)}
\newcommand{\seqjoint}{p(\theta,\data_{t};\history,\design_{t})}

\newcommand{\seqmarg}{p(\data_{t};\history,\design_{t})}

\newcommand{\updatedmodel}{p(\theta;\design,\data)}

\newcommand{\seqeig}{\eigt(\design_t ; \history)}

\newcommand{\pllm}{p_{\mathrm{LLM}}}
\newcommand{\pmodel}{p}

\newcommand{\thetallm}{\pmodel\!\left(\theta;\history\right)}
\newcommand{\likellm}{\pmodel\!\left(\data_{t};\left[\theta,\design_{t}\right]\right)}
\newcommand{\thetagen}{{\color{black}\thetallm\likellm}}
\newcommand{\yllm}{\pmodel\!\left(\data_{t};[\history,\design_{t}]\right)}
\newcommand{\postllm}{\pmodel\!\left(\theta;[\history,\design_{t},\data_{t}]\right)}
\newcommand{\ygen}{{\color{black} \yllm\postllm}}

\newcommand{\thetafilter}{p_f(\theta;\history)}
\newcommand{\seqlikellm}{\pllm\!\left(\data_{t};\left[\theta,\design_{t}\right]\right)}

\newcommand{\algcomment}[1]{\hfill\textit{\footnotesize\textcolor{gray}{// #1}}}

\makeatletter
\algrenewcommand{\algorithmiccomment}[1]{\hspace{1em}%
  \(\triangleright\)\,#1}%

\makeatother

\title{\looseness=-1\name: Intelligent Information Gathering with LLMs and Bayesian Experimental Design}

\author{%
    Deepro Choudhury \\
    University of Oxford
    \And
    Sinead Williamson \\
    Apple
    \And
    Adam Goli{\'n}ski \\
    Apple
    \And
    Ning Miao \\
    City University of Hong Kong
    \And
    Freddie Bickford Smith \\
    University of Oxford
    \And
    Michael Kirchhof \\
    Apple
    \And
    Yizhe Zhang \\
    Apple
    \And
    Tom Rainforth \\
    University of Oxford
}

\author{%
  Deepro Choudhury\inst{*},
  Sinead Williamson\inst{\dagger},
  Adam Goli{\'n}ski\inst{*,\ddagger},
  Ning Miao,
  Freddie Bickford Smith,
  Michael Kirchhof,
  Yizhe Zhang,
  Tom Rainforth
  \\[-0.2em]
  {\normalfont\small
    \inst{*}University of Somewhere \quad
    \inst{\dagger}Institute of Something \quad
    \inst{\ddagger}Company Ltd.
  }%
}

\newcommand{\inst}[1]{\ensuremath{^{#1}}}
\newcommand{\authorentry}[2]{\mbox{#1\inst{#2}}}

\author{%
  \parbox[t]{0.95\textwidth}{\centering
    \authorentry{Deepro Choudhury}{*},\allowbreak\ %
    \authorentry{Sinead Williamson}{\dagger},\allowbreak\ %
    \authorentry{Adam Goli{\'n}ski}{\dagger},\allowbreak\ %
    \authorentry{Ning Miao}{\ddagger},\allowbreak\ %
    \authorentry{Freddie Bickford Smith}{*},\allowbreak\ %
    \authorentry{Michael Kirchhof}{\dagger},\allowbreak\ %
    \authorentry{Yizhe Zhang}{\dagger},\allowbreak\ %
    \authorentry{Tom Rainforth}{*}%
    \\[-0.25em]
    {\normalfont\footnotesize
      \inst{*}University of Oxford \quad
      \inst{\dagger}Apple \quad
      \inst{\ddagger}City University of Hong Kong \quad
    }%
  }%
}

\author{%
    \authorentry{Deepro Choudhury}{*}~~\allowbreak\ %
    \authorentry{Sinead Williamson}{\dagger}~~\allowbreak\ %
    \authorentry{Adam Goli{\'n}ski}{\dagger}~~\allowbreak\ %
    \authorentry{Ning Miao}{\ddagger}\\
    \noindent
    \authorentry{\textbf{Freddie Bickford Smith}}{*}~~\allowbreak\ %
    \authorentry{\textbf{Michael Kirchhof}}{\dagger}~~\allowbreak\ %
    \authorentry{\textbf{Yizhe Zhang}}{\dagger}~~\allowbreak\ %
    \authorentry{\textbf{Tom Rainforth}}{*}%
    \\[0.5em] %
    {\normalfont
      \inst{*}University of Oxford \quad
      \inst{\dagger}Apple \quad
      \inst{\ddagger}City University of Hong Kong \quad
    }%
}

\begin{document}

\raggedbottom
\setcounter{footnote}{0}

\maketitle

\begin{abstract}
    We propose a general-purpose approach for improving the ability of large language models (LLMs) to intelligently and adaptively gather information from a user or other external source using the framework of sequential Bayesian experimental design (BED).
This enables LLMs to act as effective multi-turn conversational agents and interactively interface with external environments.
Our approach, which we call \name (Bayesian experimental design with large language models), is based on iteratively choosing questions or queries that maximize the expected information gain (EIG) with respect to a variable of interest given the responses gathered previously.
We show how this EIG can be formulated (and then estimated) in a principled way using a probabilistic model derived from the LLM's predictive distributions and provide detailed insights into key decisions in its construction and updating procedure.
We find that \name achieves substantial gains in performance across a wide range of tests based on the 20 Questions game and using the LLM to actively infer user preferences, compared to purely prompting-based design generation and other adaptive design strategies.

\end{abstract}

\section{Introduction}

Intelligent information gathering---the ability to ask the right questions at the right time---is fundamental to effective AI systems.
However, despite their many successes, LLMs currently fall short on proactively seeking out information from a user or external environment in an adaptive manner~\citep{laban2025llmslostmultiturnconversation,li2025singleturnsurveymultiturninteractions}.
For example, they have been shown to perform poorly on multi-turn guessing games~\citep{bertolazzi-etal-2023-chatgpts, zhang2024probingmultiturnplanningcapabilities}, task clarification~\citep{chi2024clarinetaugmentinglanguagemodels}, IT task automation~\citep{jha2025itbenchevaluatingaiagents} and multi-step tool use~\citep{patil2025the}.
In particular, while modern LLMs are often capable of producing coherent and insightful questions (or other external queries) in a single-turn setting, they typically struggle to appropriately tailor their questions to previously gathered responses on interactive tasks~\citep{bertolazzi-etal-2023-chatgpts,patil2025the}.

There is, therefore, a pressing need to improve the ability of LLMs to \emph{adaptively} ask questions based on previous responses, and gather information in a targeted manner.
Such capabilities are essential for a wide variety of problems, such as clarifying user intent, personalizing model behavior to a particular user, or generally acting as effective multi-turn conversational agents.
They are also critical if we want to use LLMs in data gathering tasks or as automated agents in decision-making pipelines~\citep{wu2025collabllmpassiverespondersactive}.
In turn, these capabilities are essential across domains ranging from medical diagnosis~\citep{info:doi/10.2196/59267}, troubleshooting~\citep{jha2025itbenchevaluatingaiagents}, 
preference learning~\citep{chakraborty2024maxminrlhfalignmentdiversehuman,handa2024bayesianpreferenceelicitationlanguage,ouyang2022training} and tutoring systems~\citep{kestin2025ai,liu2024socraticlm}, to conducting automated surveys~\citep{aher2023using,Jacobsen_2025,lee2024can} and AI-driven scientific inquiries~\citep{lu2024aiscientistfullyautomated,mandal2025autonomousmicroscopyexperimentslarge}.
Note that in all these problems it is not enough for the LLM to generate full sets of suitable questions up front: we need it to be able to adaptively choose questions that are tailored to the already-collected user responses.

We propose to address this challenge using the framework of \emph{sequential Bayesian experimental design} (BED;~\citealp{chaloner1995bayesian,lindley1956measure,mackay1992information,rainforth2024modern,sebastiani2000maximum}), which provides a model-based, information-theoretic mechanism for making adaptive design decisions, given a generative model of the experiment.
Specifically, we show how the problem of interactive information gathering with LLMs can be formulated as a sequential experimental design problem with a model derived from the LLM, wherein we iterate between choosing queries by maximizing expected information gain (EIG) in a variable of interest and updating our beliefs with the information from the received response.

We call our approach \name and show how its success is critically dependent on our precise model formulation, belief updating procedure and EIG estimation strategy.
In particular, we show that it is essential to formulate the model with a precise distribution pairing that does not solely rely on in-context learning to update beliefs and uses the LLM's uncertainties in the space of answers rather than the more complicated underlying hypothesis space we are trying to learn in.

\looseness=-1
Together, we find that these innovations provide substantial performance benefits over directly generating queries from the LLM and more basic approximations of the sequential BED framework.
Specifically, we first find that \name provides substantial improvements in the success rate for the 20 Questions problem across a variety of LLMs and target quantities.
On average, \name improves the success rate by 37.4 percentage points compared to direct prompting of the LLM, with the success rate more than doubling in over half of the setups considered and never decreasing.
Second, we demonstrate noticeable improvements in using the LLM for movie recommendations, showing that these benefits hold even when the LLM's predictive model differs from that of the answerer.

\section{Problem formulation and background}\label{sec:background}

There are two natural ways to improve LLMs’ ability to gather information: modifying the model itself (e.g.~via finetuning) or altering how the model is used at deployment time. 
We focus on the latter, since information-gathering tasks rarely provide task-specific data upfront (e.g.~a user’s unknown preferences), and deployment-time methods avoid the cost and difficulty of finetuning an LLM altogether and are applicable to any existing LLM. 
However, we emphasize that improvements at the model level \citep[e.g.][]{zhang2024probingmultiturnplanningcapabilities} would be complementary to our approach. 

To formalize the notion of information gathering, we need a concrete idea of what we wish to learn about. 
We denote the target quantity of interest as $\theta$, which may represent, for example, a user’s preferences, the answer to a question, or a desired piece of content.
We start with incomplete information about $\theta$, as represented by an initial \emph{belief distribution} or prior, $p(\theta)$, but can refine these beliefs by making queries, $x\in \mathcal{X}$, to the user or some other external agent and receiving responses, $y \in \mathcal{Y}$, that are informative about $\theta$.
Multiple such queries, $x_1,\dots,x_T$, can be adaptively selected in a sequential decision-making process where we iteratively choose each $x_{t}$ based on the collected history, $\history := (x_i,y_i)_{i=1}^{t-1}$. As our history grows, we will update our belief distribution to obtain $\model$ via some model updating procedure.\footnote{
We carefully distinguish between explicit probabilistic conditioning, i.e.~$p(a|b)$, and more general dependency, $p(a;b)$.
The former corresponds to the conditional distribution of an associated joint distribution, $p(a,b)$, while the latter may not. 
Here, $\history$ influences our distribution on $\theta$, but it is not derived via a joint distribution.
}
In the LLM setting, there is considerable flexibility in how $\model$ is constructed, as discussed in \Cref{sec:sBED} and \Cref{sec:justification}. 
While $\model$ need not be explicitly defined, it provides the foundation for our information-theoretic method of query selection.

For clarity of exposition, we focus on the case where the $x_t$ correspond to explicit questions asked to the user, but emphasize that the approach applies more broadly to other forms of external interaction by the LLM, such as retrieving documents or calling external functions.

\subsection{In-context updating of beliefs}

A natural and cheap way to incorporate the interaction history into the LLM is to include it in the context~\citep{brown2020language}.
If the LLM's distribution over generated text, $z \in \mathcal{Z}$, is $\pllm(z)$ given appropriate prompting, then $\pllm(z;\history)$ is an updated distribution with the previous question--response pairs in context.
From this, we can derive an updated belief distribution over~$\theta$.
Most simply, this can be done by using $\pllm(z;\history)$ to directly query about $\theta$ (e.g.~if $\theta$ is some preference, we could prompt the LLM to predict this preference).
However, as we show later, this approach often fails to appropriately incorporate the information from $\history$, leading to a belief distribution inconsistent with past observations.
This is consistent with recent work that shows that in context updating does not treat all contextual information equally \citep{kossen2024incontext,liu2023lost,zhang2024probingmultiturnplanningcapabilities}. 
In \Cref{sec:sBED}, we introduce a more robust method for deriving $p(\theta;\history)$.

\subsection{Information-theoretic experimental design}\label{sec:background-bed}

\looseness=-1
The core of the BED framework is a joint generative model, $\jointmodel$, over the target quantity, $\theta$, and the outcome, $\data$, corresponding to a design, $\design$.
Typically this is specified as a Bayesian model using a prior, $\prior$, and a likelihood, $\likelihood$.
In the general case, designs are chosen to maximize the expectation of some utility function, $U(\theta,\data,\design)$, under this model: we choose $\design^* = \argmax_{\design} \E_{\jointmodel}\left[U(\theta,\data,\design)\right]$.
The most common choice is to take $U(\theta,\data,\design) = \log \jointmodel - \log \prior \log \marginal$, 
where $\prior$ and $\marginal$ are the marginal distributions on $\theta$ and $\data$ implied by our joint model and we have assumed that our current beliefs on $\theta$ are independent of the design, $\design$.
This leads to an objective corresponding to the expected information gain (EIG) in $\theta$~\citep{lindley1956measure,lindley1972bayesian},
\begin{align}
   \eigt(\design)
    &= \mathrm{H}[\prior] - \mathbb{E}_{\marginal}[\mathrm{H}[\posterior]]
    \label{eq:bayes_eig_theta} \\
    &= \mathrm{H}[\marginal] - \mathbb{E}_{\prior}[\mathrm{H}[\likelihood]],
    \label{eq:bald} 
\end{align}
where $\mathrm{H}$ denotes the Shannon entropy (i.e. $\mathrm{H}[\prior]=-\E_{\prior}[\log \prior]$).
We can thus equivalently think of the EIG as (a) the mutual information between $\theta$ and $\data$, (b) the expected reduction in entropy over $\theta$ (i.e. information gain in $\theta$) from observing data simulated from our model, or
(c) the expected reduction in entropy over data from observing $\theta$ simulated from our prior~\citep{sebastiani2000maximum}.

Working with the EIG is highly suited to a \emph{sequential} or \emph{adaptive} design approach, generally referred to as sequential BED or Bayesian adaptive design~\citep{rainforth2024modern}.
Because the EIG is only a function of our underlying model, when we update the model as new data becomes available, our EIG design objective will naturally update as well.
Specifically, to derive the \emph{incremental} EIG~\citep{cavagnaro2010adaptive} for the $t$-th query, $\seqeig$,
we simply replace the joint in the above formulation, $p(\theta,\data;\design)$, with the updated joint,  $\seqjoint$, with all marginals and conditionals derived from this (e.g.~$p(\data;\design)$ becomes $\seqmarg$).
Here this updated joint conventionally comes from a Bayesian update of the original model.
However, in many cases, this is not practical and other non-Bayesian updates are performed instead.
For example, in active learning the update often actually corresponds to retraining the model with the new data~\citep{smith2023prediction,gal2017deep}.

\section{Sequential Bayesian experimental design with LLMs}\label{sec:method}

The sequential BED framework described in \Cref{sec:background-bed} requires two core components to be specified by the user: (a) an initial joint model, $\jointmodel$, over hypotheses, $\theta$, and outcomes, $\data$, given a design, $\design$; and (b) a procedure to derive an updated model, $\seqjoint$, after observing $\history$.
In the LLM setting, there is significant flexibility in these critical methodological decisions.
In particular, there are many ways to derive a suitable joint distribution from the LLM and its ability to learn in-context provides opportunities for update methods that go beyond standard Bayesian model updates.

\paragraph{Model construction}
A major challenge in the LLM setting is that unlike conventional probabilistic models, in general, $\pllm(\theta)\,\pllm(\data;[\theta,\design]) \neq \pllm(\data;\design)\,\pllm(\theta;[\design,\data])$.
That is, we induce a different joint distribution if we first sample $\theta$ then sample $\data$ with $\theta$ in context (which we refer to as the \emph{prior--likelihood pairing}) versus if we first sample $\data$ then sample $\theta$ with $\data$ in context (\emph{data--estimation pairing}).
Moreover, we can deviate from the distribution directly induced by the LLM on one or both variables.
The success of using BED with LLMs turns out to be critically dependent on these choices.

We delay proper discussion of this complex issue until \Cref{sec:justification}, where we will see that the preferable setup can depend on problem setting and, in particular, the relative complexity of spaces of $\theta$ and $\data$.
For now, we will focus on using the prior--likelihood pairing; we will argue in \Cref{sec:justification} that this is the advantageous setup in many practical scenarios. 
While we will generally use the LLM's directly induced distribution for the likelihood, we allow the prior to deviate from this in a problem-specific manner.
As such, our initial joint model will be
$\jointmodel = p(\theta)\pllm(\data;[\theta,\design])$.

\paragraph{Model updating}
Optimally updating the joint model in this setting requires incorporating new observations in a way that both fully captures the information from new data and is computationally tractable. 
At one extreme, we could target full Bayesian updates via approximate inference, as in classical sequential BED. 
However, this demands a prohibitively large number of LLM evaluations to accurately approximate the posterior, and it does not exploit the power of the LLM as a probabilistic generative model, where autoregressive sequential rollouts often lead to more nuanced and diverse behavior than repeated static queries.
At the other extreme, simple in-context updating, with $p(\theta;\history)=\pllm(\theta;\history)$, is cheap but, as we show later, fails to reliably capture information from new data, leading to inconsistent belief states and undermining the sequential BED approach.
As we discuss in \Cref{sec:sBED}, we therefore employ a strategy that is somewhere between the two: drawing samples in a way that utilizes $\pllm(\theta; \history)$ while encouraging diversity, then filtering out samples that are actually not compatible with $\history$ and renormalizing.
We refer to the resulting distribution as $\thetafilter$.
We do not update our likelihood model, $\pllm(\theta;\history)$; see \Cref{app:updating_likelihood} for empirical comparison of updated vs. static likelihoods in our experiments, and a discussion.

\paragraph{\name} We now introduce our specific algorithmic approach, \name.
Here, the queries will correspond to our designs, $\design$ (assumed to be in form of questions posed to the user in the following for simplicity, but could also be external function calling, document retrieval, web search, etc), and the responses received will correspond to our outcomes, $\data$.
Using the LLM to derive joint models over these outcomes and the target variable, $\theta$, given the history, $\history$, as described above, we can interleave choosing informative questions by optimizing the incremental EIG, $\seqeig$, and updating our underlying model based on the received question--response pairs.
Specifically, \name iterates over the following key steps, where $t$ indexes the current turn (see also \Cref{fig:walkthrough}).

\begin{enumerate}[label=(\Alph*),labelindent=-5pt,leftmargin=!,topsep=0pt]
    \item
    \textbf{Extract beliefs (\Cref{sec:sBED}).} Use joint model $p(\theta,\data_t;h_{t-1},\design_t) = p(\theta)\pllm(\data_t;[\theta,\design_t])$, where $h_0 = \emptyset$, for $t=1$ and $p(\theta,\data_t;h_{t-1},\design_t)=p_f(\theta;h_{t-1})\pllm(\data_t;[\theta,\design_t])$ for $t > 1$.

    \item
    \textbf{Generate candidate questions (\Cref{sec:q-gen}).} Propose a candidate set of $M$ diverse, multiple-choice questions, $\mathcal{X}^{\mathrm{cand}}$, by appropriate sampling of the LLM based on the history, $\history$.
    
    \item 
    \textbf{Estimate EIG (\Cref{sec:est}).} For each candidate, $\design_t \in \mathcal{X}^{\mathrm{cand}}$, estimate $\eigt(\design_t; \history)$.
    
    \item
    \textbf{Select and ask the best question.} Choose $x_t$ by EIG maximization and pose to the user.

    \item
    \textbf{Update the history.} Observe response, $y_t$, and update the history to $h_t = (h_{t-1},(x_t,y_t))$.
    
\end{enumerate}

\begin{figure*}[t]
    \begin{tcolorbox}[
        enhanced,
        colback=white,
        colframe=black!70,
        boxrule=0.5pt,
        arc=2pt,
        left=8pt, right=8pt, top=6pt, bottom=6pt,
        coltitle=black,
        colbacktitle=gray!15,
        fonttitle=\small\bfseries
    ]
        \small
        
        Current history: $h_2 = ((\text{``Born in the 20th century?''}, \text{``Yes''}), (\text{``Is this person male?''}, \text{``Yes''}))$.
        
        \medskip
        
        (A) \textbf{Extract beliefs (\Cref{sec:sBED})}.
        Construct hypothesis set $\Theta^{\mathrm{cand}} = \{\theta_n\}_{n=1}^N$
        by sampling candidate hypotheses, $\theta \sim p_{\mathrm{LLM}}(\theta; h_2)$, and rejecting any $\theta$ that is inconsistent with $h_2$.
        Use joint model $p(\theta,\data;h_2,\design)=p_f(\theta;h_2)\pllm(\data;[\theta,\design])$, where $p_f(\theta;h_2)$ is uniform over $\Theta^{\mathrm{cand}}$.
        
        \begin{center}
        \footnotesize
        \setlength{\tabcolsep}{8pt}
        \begin{tabular}{cccccc}
        Barack Obama & Steve Irwin & Hugh Laurie & Banksy & Elvis Presley & \ldots
        \end{tabular}
        \end{center}
        
        (B) \textbf{Generate candidate questions (\Cref{sec:q-gen}).}
        Sample $\tilde{\design}_{1:M} \sim \pllm(\tilde{\design}_{1:M};[\history, \Theta^{\mathrm{cand}}])$.
        
        \begin{center}
        \footnotesize
        \setlength{\tabcolsep}{2pt}
        \begin{tabular}{rl}
        $\tilde{\design}_{1}$: & ``Was this person born in Antarctica?'' \\
        $\tilde{\design}_{2}$: & ``Was this person born in the 19th century?'' \\
        $\tilde{\design}_{3}$: & ``Is this person an artist?'' \\
        $\tilde{\design}_{4}$: & ``Is this person European?'' \\
        $\tilde{\design}_{5}$: & ``Does this person prefer thrash metal over death metal?'' \\
        \end{tabular}
        \end{center}
        
        (C) \textbf{Estimate EIG (\Cref{sec:est}).}
        Compute 
        $\underbrace{\mathrm{H}[\hat{p}(y;[h_2,x])]}_{\text{marginal entropy}} - \underbrace{\tfrac{1}{N}\textstyle\sum_{n=1}^N \mathrm{H}[p_{\mathrm{LLM}}(y;[\theta_n,x])]}_{\text{expected conditional entropy}} \approx \mathrm{EIG}_\theta(x; h_2)$.
        
        \begin{center}
        \footnotesize
        \setlength{\tabcolsep}{4pt}
        \renewcommand{\arraystretch}{1.15}
        \begin{tabular}{clcccp{4.0cm}}
        \toprule
        & \textbf{Question} & \textbf{Marg.} & \textbf{Exp. cond.} & $\textbf{EIG}$ & \textbf{Intuition} \\
        \midrule
        $\tilde{\design}_{1}$ & Born in Antarctica? & $\approx 0$ & $\approx 0$ & $\approx 0$ & Answer ``No'' for all $\theta_n$ \\
        $\tilde{\design}_{2}$ & Born in 19th century? & $\approx 0$ & $\approx 0$ & $\approx 0$ & Redundant given $h_2$ \\
        $\tilde{\design}_{3}$ & An artist? & $0.56$ & $0.41$ & $0.15$ & Uneven split \\
        $\tilde{\design}_{4}$ & European? & $0.97$ & $0.17$ & $0.80$ & Balanced split; crisp answers \\
        $\tilde{\design}_{5}$ & Thrash vs.\ death metal? & $0.89$ & $0.88$ & $0.01$ & Uncertain even given $\theta$ \\
        \bottomrule
        \end{tabular}
        \end{center}

        \medskip

        (D) \textbf{Select and ask the best question.}
        Choose $x_3 = \tilde{\design}_{4}$ by maximizing EIG.
        Output this to the user.

        \medskip
        
        (E) \textbf{Update the history.}
        Observe answer, $y_3$.
        Set $h_3 = (h_2, (x_3, y_3))$.
    \end{tcolorbox}
    \vspace{-5pt}
    \caption{
    \looseness=-1
        \name applied to the 20 Questions game involves repeatedly constructing a belief state, generating candidate questions, estimating and maximizing EIG to select a question, and interacting with the user to gather new data.
        The contrast between $\tilde{\design}_{4}$ and $\tilde{\design}_{5}$ illustrates the benefit of using EIG with a non-deterministic likelihood: $\tilde{\design}_{5}$ has high marginal entropy (answers are uncertain), but its expected conditional entropy is equally high (answers are uncertain even given $\theta$), so nearly nothing is expected to be learned, and $\tilde{\design}_{4}$ should be favoured as a result.
        Numerical values are illustrative.
    }
    \vspace{-10pt}
    \label{fig:walkthrough}
\end{figure*}

\subsection{Prior construction and belief updating}
\label{sec:sBED}

The Savage axioms~\citep{savage1954foundations} tell us that a rational agent should update its beliefs in a Bayesian manner. 
However, doing full Bayesian updates to our model as the history grows is generally impractical for computational reasons in the LLM setting, as it requires approximate inference and this, in turn, typically requires large numbers of expensive likelihood evaluations.
Furthermore, the Savage axioms only hold if our (implied) prior truly represents our beliefs, but we find that $\pllm(\theta)$ is typically heavily overconfident on a small number of possible hypotheses and can struggle to convey the full range of possibilities even with careful prompting and a high temperature (see \Cref{fig:theta_dist}).

A natural tractable alternative is to derive our beliefs through LLM in-context updates, that is, use $\pllm(\theta;\history)$, noting that this has been shown to behave differently to Bayesian updating \citep{falck2024incontextlearninglargelanguage,kossen2024incontext}.
However, we find that even state-of-the-art LLMs such as GPT-4o~\citep{openai2024gpt4ocard} often fail to incorporate history faithfully; they regularly sample hypotheses incompatible with past observations and exhibit premature overconfidence, with both issues becoming more pronounced as $\history$ grows.
We discuss reasons for these shortfalls in \Cref{app:sBED}.

To avoid these shortfalls, we instead propose an approach that balances tractability and faithfulness.
Although we will still use $\pllm(\theta;\history)$ as the basis for deriving our belief state over $\theta$ (i.e.~our intermediate prior), we make various alterations to effectively incorporate historical information and ensure diversity. 
Our derived distribution, which we refer to as $\thetafilter$, differs from $\pllm(\theta;\history)$ in two key ways. First, we filter the generated hypotheses according to whether they are compatible with the history $\history$.
We do this by using the LLM to check the compatibility of each sampled $\theta$ with all the previous question--answer pairs in $\history$ (using $\pllm(y_i;[\theta,x_i])$ for $i \in (1, 2, \ldots, t-1)$) and then rejecting that sample if an incompatibility is found.
Specifically, a sample is rejected if the likelihood of an observed answer falls below a predefined threshold, chosen to balance robustness to model uncertainty against the need to enforce strict historical coherence.
To reduce the computational cost of generating and evaluating hypotheses, we further include a \emph{hypothesis-retention mechanism}: any hypotheses from the previous turn which remain consistent with the most recent question and observation are retained in the hypothesis set %
without regeneration.

Second, we make a number of modifications to promote diversity. Rather than generate candidates independently, we prompt the LLM to generate batches of candidates using a prompt encouraging diversity. After filtering these candidates as above and removing duplicates, we then impose a uniform distribution. Details of our exact setup for $\thetafilter$ are given in~\Cref{sec:candidate_hypotheses}.

\subsection{Generating candidate questions}\label{sec:q-gen}

As it is not computationally feasible to directly optimize over the space of possible questions, we rely on using the LLM to propose diverse candidate questions, $\mathcal{X}^{\mathrm{cand}}$, then select the best question from these.
We consider two specific approaches.
(a)~\emph{Unconstrained generation:} given $\history$, the LLM is simply asked to propose new questions by sampling from $\pllm(\design_{t};\history)$ with appropriate prompting. 
(b)~\emph{Conditional generation:} the LLM is given both $\history$ \textit{and} a generated set of hypotheses, $\Theta^{\mathrm{cand}} = \{\theta_n\}_{n=1}^N$, such that we sample from $\pllm(\design_{t};[\history, \Theta^{\mathrm{cand}}])$; specifically, the LLM is prompted for questions that ``slice'' the hypothesis pool into roughly balanced subsets.

For both strategies, we sample $M$ questions jointly with a relatively high temperature to encourage diversity. Conditional generation allows us to ``guide'' the LLM to propose highly informative questions. 
However, it risks overfitting to $\Theta^{\mathrm{cand}}$. 
In practice, we find it is effective for discrete spaces (\Cref{sec:20q}), 
but less so for spaces with complex, overlapping hypotheses (\Cref{sec:personalisation_results}). 
We restrict questions to multiple-choice format to simplify uncertainty quantification (see \Cref{sec:justification}).

\subsection{Estimating EIG for each question}\label{sec:est}

To estimate the EIG based on Equation~\ref{eq:bald} for a given question $\design_{t}$, we use the following Rao-Blackwellized estimator based on the LLM's predictive distribution:
\begin{align}
    \label{eq:est-rb}
    \begin{split}
    \seqeig \approx& \textstyle 
    \frac{1}{N} \sum_{n=1}^N \sum_{\data_{t} \in \mathcal{Y}} \pllm(\data_{t} ; [\theta_n, \design_{t}]) \log \pllm(\data_{t} ; [\theta_n, \design_{t}]) \\
    &-\textstyle \sum_{\data_{t} \in \mathcal{Y}} \hat{p}(\data_{t} ; [\history,\design_{t}]) \log \hat{p}(\data_{t} ; [\history,\design_{t}]),
    \end{split}
\end{align}
\looseness=-1
where $\hat{p}(\data_{t} ; [\history,\design_{t}]) \!:=\! \frac{1}{N} \sum_{n=1}^N \pllm(\data_{t} ; [\theta_n, \design_{t}])$ and $\theta_n \!\sim\! \thetafilter$ (see \Cref{sec:sBED}).
This estimator has been used in other BED contexts~\citep{gal2017deep, rainforth2017automating}.
Note that the samples do not need to be independent for this estimator to converge, provided they satisfy some appropriate form of ergodicity or decaying correlation~(see, e.g.,~\cite{billingsley2013convergence}). 
When constructing this estimator, we compute the $\pllm(\data_{t} ; [\theta_n, \design_{t}])$ terms using the LLM's logits whenever possible.
By the Rao-Blackwell theorem, 
this always produces lower variance than purely sample-based estimators~\citep{rao1945information}, like those employed in~\cite{hu2024uncertaintythoughtsuncertaintyawareplanning} and \citet{kobalczyk2025active}.

\paragraph{Avoiding deterministic likelihood assumptions}\label{sec:pred_ent}
Previous attempts to apply information criteria to choosing queries in LLMs have generally assumed responses are deterministic given $(\theta,\design_t)$~\citep{cooper2025curious,kobalczyk2025active,hu2024uncertaintythoughtsuncertaintyawareplanning,mazzaccara2024learningaskinformativequestions,piriyakulkij2023active}.
Under this assumption, the EIG simplifies to the marginal predictive entropy, 
$\mathrm{H}[p(\data_{t};\design_{t},\history)]$.

This is problematic as, in practice, the expected likelihood entropy will vary with $\design_t$. 
In general, $\E_{p(\theta;\history)}[\mathrm{H}[p(\data_{t}|\theta;\design_{t},\history)]]$ measures how certainly the question can be answered once $\theta$ is known.
Including it in our objective is essential in avoiding questions that are irrelevant, ambiguous, unclear, or simply unhelpful in our quest to learn about $\theta$.
\Cref{fig:walkthrough} illustrates this concretely---question $\tilde{x}_4$ has high marginal entropy and low conditional entropy (answers are crisp given $\theta$), yielding high EIG, while $\tilde{x}_5$ has equally high marginal entropy but also high conditional entropy (answers are noisy \emph{even given} $\theta$), so the terms cancel and nearly nothing is learned.
An entropy-only method would score both similarly and risk wasting a turn; the same failure mode causes the Entropy ablation to provide no improvement over \baseline in our preference elicitation experiments (\Cref{sec:personalisation_results}).
Given that both terms of~\Cref{eq:est-rb} are computed from the same LLM likelihood evaluations, retaining the full EIG requires no additional calls and provides no computational savings. Thus, we advise against making deterministic likelihood assumptions.

{\color{red}

}

\section{On the specification of $\seqjoint$, and its implications}
\label{sec:justification}

As we described in \textsection\ref{sec:method}, successfully applying sequential BED in the LLM setting hinges upon how we specify, and update, the joint distribution, $\seqjoint$. 
In particular, as previously highlighted, there are two distinct ways to derive the joint model from our LLM: using a \emph{prior--likelihood pairing}, $\thetagen$, or a \emph{data--estimation pairing}, $\ygen$.
The first construction mirrors deriving our beliefs about $\theta$ from a \emph{conventional} Bayesian posterior with a concrete prior and likelihood derived (at least partially) from the LLM, whereas the second has analogies to a \emph{marginal-posterior} approach~\citep{fong2023martingale, falck2024incontextlearninglargelanguage} in that it that samples hypothetical data and draws inferences on $\theta$ given hypothetical data using in-context learning.
In our outlined \name approach, we adopted a {prior--likelihood} pairing. Below, we justify this decision and also discuss certain settings where the data--estimation setup might be preferable instead.

\textbf{Modeling flexibility}~~
The most obvious relative merits of the prior--likelihood and data--estimation pairings are in the flexibility in how each term is chosen. The {prior--likelihood} pairing gives us greater flexibility to construct a prior set of beliefs over $\theta$ that is distinct to the LLM's internal beliefs, as it allows us to directly control this prior by changing $\thetallm$, whereas the prior is only implicitly defined in the data--estimation pairing. 
In \textsection\ref{sec:sBED} we exploited this flexibility through our definition of $\thetafilter$.
On the other hand, the {data--estimation} pairing could provide some beneficial flexibility in specifying how the data itself is simulated through changing $\yllm$, which
could, for example, be useful when we have access to external data simulators.

\textbf{Faithfulness of conditional distributions}~~
While deviations from relying on direct LLM predictions are also in principle possible for the conditional models $\likellm$ and $\postllm$, in practice, these will typically be more difficult and expensive to implement.
This is first because these conditionals need to be instantiated for each sampled instance of the conditional variable ($\theta$ and $\data_{t}$ respectively), rather than just needing us to set up a single marginal distribution.
Second, to construct estimators for Equations~\eqref{eq:bayes_eig_theta} and \eqref{eq:bald}, we require access to concrete \emph{probabilities} for the conditional distributions (in order to calculate entropies), whereas we only needed to draw samples for the marginal distributions (in order to approximate expectations). 
As such, the conditionals need to be explicit distributions, or at least ones where the probability can be cheaply estimated, so they are more difficult to define through the output of some algorithmic procedure, especially in large spaces.

When considering the conditional distributions, the decisive question on the relative merit of the two formulations is which conditional factor we are willing to trust the LLM to supply as a \emph{full probability distribution}. Critically, we rely on how the LLM captures uncertainty in this {full distribution}---including, for example, tail behavior---not merely the fidelity of typical samples; the marginal factors, by contrast, only need to be sampled from. 
If we accept the LLM's direct predictive distribution for $\likellm$, then we are basing our notion of uncertainty around (and will need to calculate) $\mathrm{H}[\seqlikellm]$, and if instead we place more faith in the LLM's internal distribution for $\postllm$, then we are basing our uncertainty around $\mathrm{H}[\pllm(\theta;[\history,\design_{t+1},\data_{t+1}])]$. 
In essence, the choice between prior--likelihood and data--estimation pairings thus comes down to whether we believe the LLM will produce a more appropriate conditional uncertainty over $\theta$ or $\data$, along with our ability to numerically estimate this uncertainty cheaply.

\looseness=-1
This difference becomes particularly noticeable when the complexities of the spaces of $\theta$ and $\data$ differ significantly.
Our ability to draw sensible samples of either will generally be quite robust to these spaces being complex or high-dimensional; this is where LLMs tend to thrive, effectively generating highly complex outputs in an autoregressive manner.
However, evaluating the entropy of a distribution becomes dramatically harder as the dimensionality or complexity increases~\citep{acharya2019estimating, paninski2003estimation}, and the entropy of the predictive distribution of an LLM in such cases will \emph{not} typically provide a sensible measure of uncertainty \citep{kadavath2022languagemodelsmostlyknow, desai-durrett-2020-calibration}.
As such, the decision between joint formulations should predominantly be based on the complexity of the space of $\theta$ versus that of $\data$: \emph{we should generally favor the prior--likelihood formulation if $\theta$ is more complex and the data--estimation formulation if $\data$ is more complex}.
For the problems that we consider, the space of $\data$ is less complex than that of $\theta$, indicating we should, in general, use the prior--likelihood formulation.
However, in cases where this is not true, the data--estimation formulation may be preferable instead,
see \Cref{app:faithfulness_cont} for further discussion on this and on the choice of $\theta$.

\looseness=-1
\textbf{Extracting the belief state}~~
A further advantage of the prior--likelihood construction is that our belief state on $\theta$ can be extracted directly as $\thetafilter$. With the data--estimation construction, we would have to estimate the marginal on $\theta$  by integrating $\ygen$ over the synthetic response $\data_t$.
Direct access to $p(\theta; h_{t-1})$ is also important to ensure that our current belief state is independent of the next question $x_{t}$, which is both intuitively desirable and theoretically required to be a valid BED approach~\citep{lindley1972bayesian}; data--estimation formulations will generally violate this.

\vspace{-4pt}
\section{Related work}
\vspace{-4pt}

Several works have explored the baseline ability of LLMs to rapidly learn about a parameter of interest by asking questions \citep{zhang2024probingmultiturnplanningcapabilities,li2025eliciting}---effectively our \emph{\baseline} baseline in \textsection\ref{sec:experiments}. 
While these works demonstrate some ability to adaptively construct information-seeking questions, they often fail to extract important information \citep{li2025questbench}.

\looseness=-1
Some works have further specifically attempted to choose questions based on model-based uncertainty criteria~\citep{piriyakulkij2023active,hu2024uncertaintythoughtsuncertaintyawareplanning,kobalczyk2025active,mazzaccara2024learningaskinformativequestions,cooper2025curious}. 
None of these works provide the same careful consideration of how the underlying joint model should be formulated, which underpins our own work, 
and they all assume deterministic likelihood models that mean their objectives correspond to a sample-based estimate of marginal predictive entropy in practice, as explained in~\Cref{sec:est}.
In general, these previous works have also required restrictions on the space of allowable hypotheses, $\theta$, and typically require additional assumptions and/or approximations.
More extensive discussion of related work is given in \Cref{app:extended_related}.

\vspace{-4pt}
\section{Experiments}
\label{sec:experiments}
\vspace{-4pt}

We now assess how well \name and alternative information-gathering approaches work in two practical scenarios: 20 Questions, a game in which the player has to guess a target entity and can ask up to 20 yes-no questions about the entity; and preference elicitation, a task in which the agent has to predict a user's preference profile and can ask five multiple-choice questions to the user.

\looseness=-1
\paragraph{Answerer}
We produce answers to the \emph{questioner} LLM's questions using a separate \emph{answerer} LLM. 
The answerer is provided with a ground-truth $\theta^*$ (a target entity in 20 Questions or a user profile in preference elicitation) and processes individual questions from the questioner without access to any of the questioner's context (i.e. $h_{t-1}$ and the questioner's prompts).
We test two questioner--answerer setups, where the two are served by separate instances of the same LLM, or two different LLMs.
The latter scenario is important 
because in practice, 
the answerer will often follow a different distribution than the questioner’s
internal model for reasoning about responses, thereby forming a model misspecification.

\looseness=-1
\paragraph{Baselines}
We compare \name against three baselines.
\emph{\baseline} prompts the questioner to directly generate an informative next question, without explicit hypothesis generation or a data-acquisition objective, and then sampling the question with temperature $T\!=\!1$; this was explored by \citet{zhang2024probingmultiturnplanningcapabilities}.
\emph{CoT} augments \baseline with a ReAct-style \citep{yao2023react} chain-of-thought: the LLM first produces a thought (what has been established, what information would help most) then an action (the question). This tests whether structured reasoning alone can close the gap to \name.
\emph{Split} chooses the question that most equally splits a sampled set of hypotheses, $\Theta^{\mathrm{cand}}$, which corresponds to maximizing the marginal predictive entropy, $\mathrm{H}[p(\data_{t};\design_{t},\history)]$, in a model with a deterministic likelihood.
As such, the methods of \citet{cooper2025curious}, \citet{hu2024uncertaintythoughtsuncertaintyawareplanning}, \citet{kobalczyk2025active}, \citet{mazzaccara2024learningaskinformativequestions} and \citet{piriyakulkij2023active} can all be viewed as variants of this objective.
To the best of our knowledge, Split represented the previous state-of-the-art method for 20 Questions.
We note that our own specific Split baseline implementation, which uses \name's filtering mechanism, also achieves dramatically better results than reported by, for example, \citet{kobalczyk2025active}, so this constitutes a strong baseline relative to previous work.
On top of this, our implementation of \baseline appears to significantly improve over that of \citet{zhang2024probingmultiturnplanningcapabilities}. 
Additional evaluations and full plots are in \Cref{app:experimental_results}.

\vspace{-4pt}
\subsection{20 Questions}\label{sec:20q}
\vspace{-4pt}

\looseness=-1
We consider three sets of 20 Questions problems: Animals, Celebrities, and Things (See \Cref{app:20q_datasets}). %
Each problem set comprises 100 target entities, $\{\theta^*_i\}_{i=1}^{100}$.
The space of possible $\theta$ is large and not explicitly defined or restricted: we do not tell the LLM this set of target entities, so the space is bounded only by what the LLM can generate; by comparison, many previous works have relied on restricted spaces for $\theta$~\citep{chan2025conformal,hu2024uncertaintythoughtsuncertaintyawareplanning,piriyakulkij2023active,wang2025adaptiveelicitationlatentinformation}.

\begin{figure}[t]
    \centering
    \includegraphics[trim={0 0.2cm 0 0.2cm},clip,width=\linewidth]{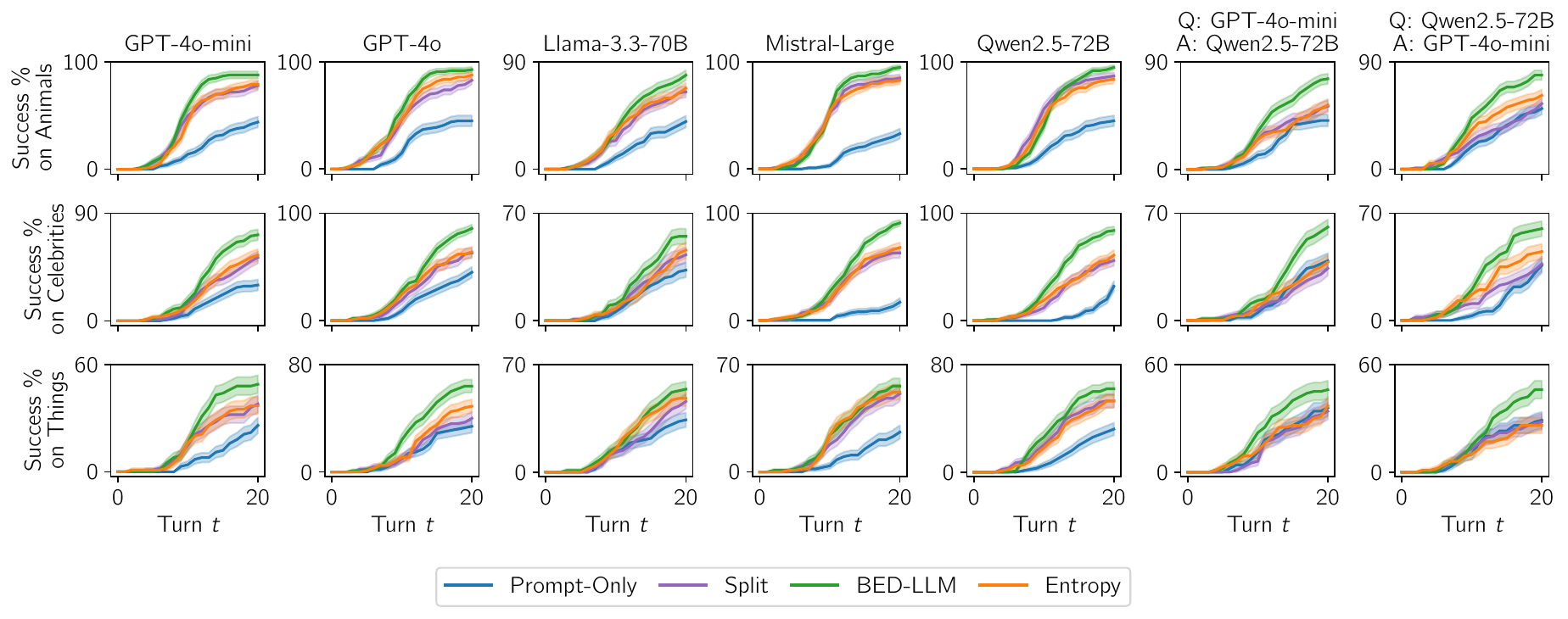}
    \vspace{-16pt}
    \caption{
        Success rate on 20 Questions: mean $\pm$ standard error across 100 targets per dataset.
        \vspace{-12pt}
    }
    \label{fig:20q_same_different_models}
\end{figure}

\begin{table}[tb]
  \centering
  \begingroup
  \setlength{\tabcolsep}{3pt}
  \renewcommand{\arraystretch}{1.05}
  \footnotesize
  \resizebox{\linewidth}{!}{%
  \begin{tabular}{llcccccccc}
    \toprule
    \multicolumn{2}{c}{} & \multicolumn{8}{c}{\textbf{Success Rate (\%)}} \\
    \cmidrule(l){3-10}
    & \textbf{Model} & {\textbf{\baseline}} & {\textbf{Split}} & {\textbf{CoT}} & {\textbf{\name}} & {\textbf{Entropy}} & {\textbf{Data--Est.}} & {\textbf{ICL Beliefs}} & {\textbf{Impl. Max.}} \\
    \midrule
    \multirow{6}{*}{\rotatebox{90}{Animals}~}
      & GPT-4o-mini   & 44\tiny{$\pm$5.0} & 78\tiny{$\pm$4.2} & 55\tiny{$\pm$5.0} & \textbf{88}\tiny{$\pm$3.3} & 79\tiny{$\pm$4.1} & 64\tiny{$\pm$4.8} & 18\tiny{$\pm$3.9} & 47\tiny{$\pm$5.0} \\
      & GPT-4o        & 45\tiny{$\pm$5.0} & 83\tiny{$\pm$3.8} & 62\tiny{$\pm$4.9} & \textbf{93}\tiny{$\pm$2.6} & 88\tiny{$\pm$3.3} & 70\tiny{$\pm$4.6} & 25\tiny{$\pm$4.4} & 70\tiny{$\pm$4.6} \\
      & Llama-3.1-8B  & 8\tiny{$\pm$2.7}  & 49\tiny{$\pm$5.0} & 19\tiny{$\pm$3.9} & \textbf{63}\tiny{$\pm$4.9} & 54\tiny{$\pm$5.0} & 38\tiny{$\pm$4.9} & 25\tiny{$\pm$4.4} & 16\tiny{$\pm$3.7} \\
      & Llama-3.3-70B & 40\tiny{$\pm$4.9} & 65\tiny{$\pm$4.8} & 40\tiny{$\pm$4.9} & \textbf{79}\tiny{$\pm$4.1} & 68\tiny{$\pm$4.7} & 40\tiny{$\pm$4.9} & 33\tiny{$\pm$4.7} & 54\tiny{$\pm$5.0} \\
      & Mistral-Large & 33\tiny{$\pm$4.7} & 85\tiny{$\pm$3.6} & 35\tiny{$\pm$4.8} & \textbf{95}\tiny{$\pm$2.2} & 83\tiny{$\pm$3.8} & 83\tiny{$\pm$3.8} & 53\tiny{$\pm$5.0} & 53\tiny{$\pm$5.0} \\
      & Qwen2.5-72B   & 45\tiny{$\pm$5.0} & 87\tiny{$\pm$3.4} & 51\tiny{$\pm$5.0} & \textbf{95}\tiny{$\pm$2.2} & 85\tiny{$\pm$3.6} & 68\tiny{$\pm$4.7} & 46\tiny{$\pm$5.0} & 61\tiny{$\pm$4.9} \\
    \midrule
    \multirow{6}{*}{\rotatebox{90}{Celebrities}~}
      & GPT-4o-mini   & 30\tiny{$\pm$4.6} & 53\tiny{$\pm$5.0} & 42\tiny{$\pm$5.0} & \textbf{72}\tiny{$\pm$4.5} & 55\tiny{$\pm$5.0} & 32\tiny{$\pm$4.7} & 16\tiny{$\pm$3.7} & 31\tiny{$\pm$4.7} \\
      & GPT-4o        & 45\tiny{$\pm$5.0} & 63\tiny{$\pm$4.9} & 63\tiny{$\pm$4.9} & \textbf{86}\tiny{$\pm$3.5} & 64\tiny{$\pm$4.8} & 55\tiny{$\pm$5.0} & 52\tiny{$\pm$5.0} & 50\tiny{$\pm$5.0} \\
      & Llama-3.1-8B  & 10\tiny{$\pm$3.0} & 35\tiny{$\pm$4.8} & 16\tiny{$\pm$3.7} & \textbf{58}\tiny{$\pm$5.0} & 36\tiny{$\pm$4.8} & 19\tiny{$\pm$3.9} & 24\tiny{$\pm$4.3} & 19\tiny{$\pm$3.9} \\
      & Llama-3.3-70B & 33\tiny{$\pm$4.7} & 43\tiny{$\pm$5.0} & 36\tiny{$\pm$4.8} & \textbf{55}\tiny{$\pm$5.0} & 46\tiny{$\pm$5.0} & 26\tiny{$\pm$4.4} & 27\tiny{$\pm$4.5} & 37\tiny{$\pm$4.9} \\
      & Mistral-Large & 19\tiny{$\pm$4.0} & 63\tiny{$\pm$4.9} & 42\tiny{$\pm$5.0} & \textbf{91}\tiny{$\pm$2.9} & 68\tiny{$\pm$4.7} & 66\tiny{$\pm$4.8} & 31\tiny{$\pm$4.7} & 36\tiny{$\pm$4.8} \\
      & Qwen2.5-72B   & 32\tiny{$\pm$4.7} & 56\tiny{$\pm$5.0} & 48\tiny{$\pm$5.0} & \textbf{84}\tiny{$\pm$3.7} & 59\tiny{$\pm$4.9} & 34\tiny{$\pm$4.8} & 26\tiny{$\pm$4.4} & 39\tiny{$\pm$4.9} \\
    \midrule
    \multirow{6}{*}{\rotatebox{90}{Things}~}
      & GPT-4o-mini   & 26\tiny{$\pm$4.4} & 38\tiny{$\pm$4.9} & 33\tiny{$\pm$4.7} & \textbf{49}\tiny{$\pm$5.0} & 37\tiny{$\pm$4.9} & 26\tiny{$\pm$4.4} & 19\tiny{$\pm$4.0} & 25\tiny{$\pm$4.4} \\
      & GPT-4o        & 34\tiny{$\pm$4.8} & 40\tiny{$\pm$4.9} & 49\tiny{$\pm$5.0} & \textbf{64}\tiny{$\pm$4.8} & 49\tiny{$\pm$5.0} & 26\tiny{$\pm$4.4} & 19\tiny{$\pm$3.9} & 42\tiny{$\pm$5.0} \\
      & Llama-3.1-8B  & 10\tiny{$\pm$3.0} & 12\tiny{$\pm$3.3} & 10\tiny{$\pm$3.0} & \textbf{26}\tiny{$\pm$4.4} & 15\tiny{$\pm$3.6} & 9\tiny{$\pm$2.9} & 11\tiny{$\pm$3.1} & 10\tiny{$\pm$3.0} \\
      & Llama-3.3-70B & 34\tiny{$\pm$4.8} & 46\tiny{$\pm$5.0} & 35\tiny{$\pm$4.8} & \textbf{55}\tiny{$\pm$5.0} & 48\tiny{$\pm$5.0} & 19\tiny{$\pm$3.9} & 15\tiny{$\pm$3.6} & 34\tiny{$\pm$4.8} \\
      & Mistral-Large & 26\tiny{$\pm$4.4} & 51\tiny{$\pm$5.0} & 29\tiny{$\pm$4.6} & \textbf{58}\tiny{$\pm$5.0} & 52\tiny{$\pm$5.0} & 46\tiny{$\pm$5.0} & 19\tiny{$\pm$3.9} & 30\tiny{$\pm$4.6} \\
      & Qwen2.5-72B   & 32\tiny{$\pm$4.7} & 51\tiny{$\pm$5.0} & 46\tiny{$\pm$5.0} & \textbf{62}\tiny{$\pm$4.9} & 51\tiny{$\pm$5.0} & 39\tiny{$\pm$4.9} & 24\tiny{$\pm$4.3} & 40\tiny{$\pm$4.9} \\
    \bottomrule
  \end{tabular}%
  }%
  \endgroup
  \caption{
    Success rate (\%) for 20 Questions at the end of the game.
    Best result in bold.
    $\pm$ numbers show the standard error of the mean 
    estimated using $\sqrt{p(1-p)/(n-1)}$ where $p$ is the success percentage and $n$ is the number of datapoints. 
    This estimator is positively biased and thus conservative.
  }
  \label{tab:20q}
\end{table}

\begin{table}[t]
\centering

\label{tab:method-summary}
\small
\begin{tabular}{lccc}
\toprule
\textbf{Method} & \textbf{Joint model} & \textbf{Objective} & \textbf{Belief updates} \\
\midrule
\textbf{\name} & Prior--likelihood & Full EIG (Eq.~\ref{eq:est-rb}) & Filtered ($p_f$) \\
\midrule
\baseline & \cellcolor{gray!20} None (implicit LLM) & \cellcolor{gray!20} None (implicit LLM) & \cellcolor{gray!20} Raw ICL ($\pllm$) \\
Split & \cellcolor{gray!20} Deterministic likelihood & \cellcolor{gray!20} Pred.\ entropy & Filtered ($p_f$) \\
\midrule
Entropy & Prior--likelihood & \cellcolor{gray!20} Pred.\ entropy & Filtered ($p_f$) \\
Data--Estimation & \cellcolor{gray!20} Data--estimation & Full EIG & Filtered ($p_f$) \\
ICL Beliefs & Prior--likelihood & Full EIG & \cellcolor{gray!20} Raw ICL ($\pllm$) \\
Implicit Max.\ & Prior--likelihood & \cellcolor{gray!20} LLM judgment & Filtered ($p_f$) \\
\bottomrule
\end{tabular}
\caption{
\looseness=-1 
Summary of how each method relates to BED-LLM's three core algorithmic components. Each ablation (bottom section) modifies exactly one component of BED-LLM, highlighted in \colorbox{gray!20}{grey}.}
\vspace{-10pt}
\end{table}

To evaluate performance, at each turn, $t \in (0, 1, \ldots, 20)$, we extract $\theta_i^t$ from $p_f(\theta_i; h_t)$ using greedy decoding and we compute the success rate as the mean across $i$ of $\mathbb{I}(\theta_i^t = \theta_i^*)$.
These evaluation guesses are not part of the questioner algorithm itself and are not included in $\history$.
In line with the original rules of the game, we also introduce an explicit mechanism for the questioner to guess the answer as one of its 20 questions: if the set of filtered hypotheses collapses to a single candidate, 
the questioner asks ``Is it $\langle$\texttt{item}$\rangle$?''.
A correct guess ends the game; otherwise the negative response is added to $\history$ and counted towards the budget. See \Cref{app:20q_details} for further experimental details.

\paragraph{\name improves over \baseline and Split baselines}
Our results in \Cref{fig:20q_same_different_models} and  \Cref{tab:20q}  show \name significantly outperforming both baselines across all problems and LLMs.
Particularly notable is that \name's final success rate is typically more than double that of \baseline, highlighting 
the substantial gains that can be achieved by using explicit EIG maximization.

\paragraph{\name ablations}
We further evaluate four ablations of \name to isolate the contribution of each of its core components;
full descriptions are in \Cref{app:ablations}.
\emph{Entropy} replaces the full EIG objective with the marginal predictive entropy, $\mathrm{H}[p(\data_{t};\design_{t},\history)]$; it is similar to the  \textbf{Split} baseline, but uses \name's likelihood model instead of a deterministic one.
\emph{Data--Estimation} swaps \name's prior--likelihood joint model for a data--estimation pairing (see \Cref{sec:data_estimation_method}), testing the importance of \name measuring uncertainty in $y$ space instead of $\theta$ space (see~\Cref{sec:method}).
\emph{ICL Beliefs} omits our filtering procedure in the belief extraction (see~\Cref{sec:sBED}), just using the, often incoherent,
raw in-context beliefs $\pllm(\theta;\history)$ instead of $\thetafilter$.
Finally, \emph{Implicit Maximization} replaces explicit EIG estimation with LLM judgment, drawing on the Tree of Thoughts \citep[ToT;][]{yao2023tree} intuition of reasoning over hypothetical futures: the model is presented with the same candidate questions as \name, but is simply prompted to internally reason and select the question it judges most informative, providing a lightweight alternative to explicit EIG estimation.

In \Cref{tab:20q} we see that \name comfortably outperforms all alternative approaches.
Notably, Entropy provides the strongest alternative, with performance slightly better than Split.
The fact that Entropy performs much more similarly to Split than \name shows that the use of a non-deterministic likelihood is beneficial predominantly in allowing us to target a proper EIG, rather than due to changes in 
the marginal predictive entropy itself.
The improvement of Implicit Maximization over Prompt-only is also notable, given how cheap it is to run (see~\Cref{fig:wall-clock} for run time information).

\paragraph{Prior--likelihood outperforms data--estimation}
Our analysis in \textsection\ref{sec:justification} 
is
validated by our results: \name's prior-likelihood approach substantially outperforms Data--Estimation.
Data--Estimation still outperforms \baseline, 
but interestingly it performs worse than Entropy, highlighting the importance of estimating uncertainty in the $\data$ space instead of $\theta$ space here.
These findings reinforce our claim that the choice of joint-model factorization is a critical algorithmic decision.

\paragraph{Rejection sampling and explicit EIG maximization are key}
We also see how we produce our beliefs over $\theta$ matters: deriving beliefs using simple in-context learning, as in ICL Beliefs, lead to massive performance drops.
Further, while \name's routines for sampling candidate questions and hypotheses are crucial, they alone are not sufficient: passing the samples to an LLM and prompting it to select the highest-EIG question, as in Implicit Maximization, works much less well.

\paragraph{\name is robust to questioner--answerer mismatch}
Our results in \Cref{fig:20q_same_different_models} demonstrate that 
the benefit of 
\name 
persists even under model misspecification.
This is important for 
applicability to real-world users, whose responses will follow a different distribution than the questioner LLM.

\subsection{Preference elicitation}
\label{sec:personalisation_results}

Unlike 20 Questions, in which $\theta$ is a concrete entity and most reasonable questions have clear answers, many real-world information-gathering tasks involve more abstract targets and less predictable data generation.
A key example is learning user preferences, where it may be difficult to explicitly define a concrete closed set of possible $\theta$, or for the LLM to develop appropriate uncertainty estimates.
To study such a scenario, we evaluate \name on inferring users' film preferences.
Here the target $\theta$ is somewhat abstract, and we have more flexibility in how we define it in our joint model.
Our chosen setup is to define $\theta$ to be a user profile, namely a paragraph of text describing the user's film tastes with our answerer model prompted to emulate a user with a given profile; see \Cref{app:preference_details} for full details.
We consider 200 ground-truth profiles, $\{\theta^*_i\}_{i=1}^{200}$, which, as with 20 Questions, are never revealed to the questioner.
Because Split is not applicable as a baseline here (a deterministic likelihood assumption is clearly unreasonable), we benchmark with the similar Entropy approach instead.
We also note that data--estimation setup is completely unviable here due to the large $\theta$ space.

At each turn, $t \in (0, 1, \ldots, 5)$, the questioner uses $\history$ to generate ten film recommendations.
This list is then rated in its fit to the user profile using an LLM-as-judge setup \citep{trivedi2024selfrationalizationimprovesllmfinegrained,zhu2025judgelmfinetunedlargelanguage}.
Specifically, the answerer scores each film on a scale of 1 to 5 (in 0.5 increments), based on how well the film aligns with $\theta^*$; this score is output together with a brief justification to increase reliability.
The films' scores are not included in $\history$.

\begin{figure}[t]
    \includegraphics[trim={0 0.2cm 0 0.2cm},clip,width=\linewidth]{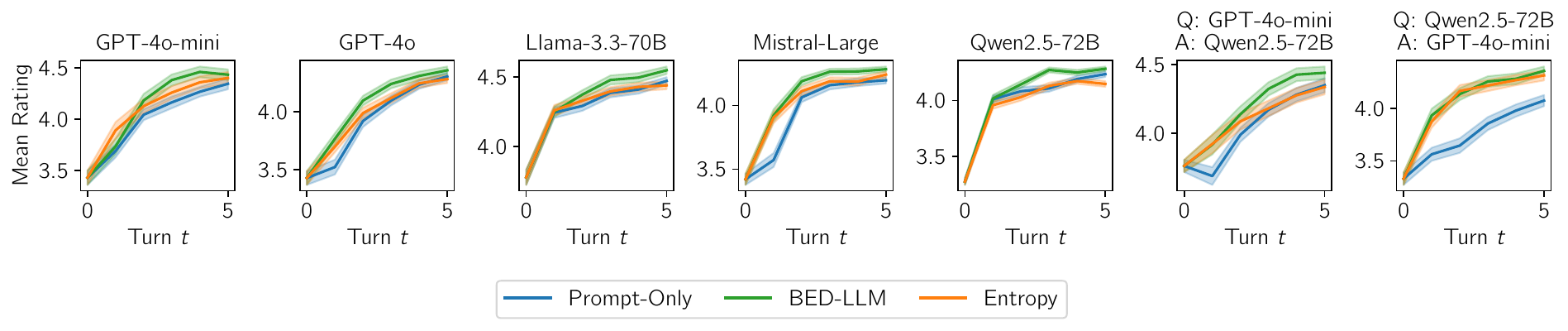}
    \vspace{-15pt}
    \caption{
        Mean rating across 10 film recommendations: mean $\pm$ standard error across 200 users.
    }
    \vspace{-10pt}
    \label{fig:preferences_same_different_models}
\end{figure}

\looseness=-1
Our results in \Cref{fig:preferences_same_different_models} show that, while \baseline seems to be a stronger baseline in this preference-elicitation scenario than for 20 Questions, \name is still able to provide a boost over both it and Entropy, producing higher-rated film recommendations.
\name's benefit is most clear in scenarios where the questioner belongs to a different model class to the answerer.

\section{Conclusion}

\looseness=-1
In this work, we have shown how to effectively apply the framework of sequential Bayesian experimental design (BED) to the problem of interactive information gathering with LLMs.
Specifically, we have introduced \name, which provides a specific, information-theoretic, sequential BED approach that makes a variety of carefully justified design choices in the joint-model factorization, belief updating, and EIG estimation. 
Particularly central to \name is the prior--likelihood pairing with filtering of hypotheses for consistency with the history.
\name is notably the first work that uses both this prior--likelihood pairing without making a deterministic likelihood assumption that causes the EIG to simply to just marginal predictive entropy.
Together, these innovations lead to substantial performance improvements compared to previous approaches.
The results thus confirm that principled EIG-driven strategies can yield substantial gains for interactive, multi-turn information gathering problems.

\newpage

\section*{Acknowledgments}
DC is supported by the EPSRC CDT in Statistics and Machine Learning (EP/Y034813/1), an Oxford-Radcliffe Scholarship and an Oxford AI Fast Exploration grant.
FBS is supported by the EPSRC Probabilistic AI Hub (EP/Y028783/1).
TR is supported by the UK EPSRC grant EP/Y037200/1.

\bibliography{main.bib}

\begin{thebibliography}{68}
\providecommand{\natexlab}[1]{#1}
\providecommand{\url}[1]{\texttt{#1}}
\expandafter\ifx\csname urlstyle\endcsname\relax
  \providecommand{\doi}[1]{doi: #1}\else
  \providecommand{\doi}{doi: \begingroup \urlstyle{rm}\Url}\fi

\bibitem[Acharya et~al.(2019)Acharya, Bhadane, Indyk, and Sun]{acharya2019estimating}
Jayadev Acharya, Sourbh Bhadane, Piotr Indyk, and Ziteng Sun.
\newblock Estimating entropy of distributions in constant space.
\newblock In \emph{Advances in Neural Information Processing Systems}, 2019.

\bibitem[Aher et~al.(2023)Aher, Arriaga, and Kalai]{aher2023using}
Gati~V Aher, Rosa~I Arriaga, and Adam~Tauman Kalai.
\newblock Using large language models to simulate multiple humans and replicate human subject studies.
\newblock In \emph{International Conference on Machine Learning}, 2023.

\bibitem[Andukuri et~al.(2024)Andukuri, Fr{\"a}nken, Gerstenberg, and Goodman]{andukuri2024star}
Chinmaya Andukuri, Jan-Philipp Fr{\"a}nken, Tobias Gerstenberg, and Noah~D Goodman.
\newblock Star-gate: Teaching language models to ask clarifying questions.
\newblock In \emph{Conference on Language Modeling}, 2024.

\bibitem[Bertolazzi et~al.(2023)Bertolazzi, Mazzaccara, Merlo, and Bernardi]{bertolazzi-etal-2023-chatgpts}
Leonardo Bertolazzi, Davide Mazzaccara, Filippo Merlo, and Raffaella Bernardi.
\newblock {C}hat{GPT}{'}s information seeking strategy: Insights from the 20-questions game.
\newblock In \emph{International Natural Language Generation Conference}, 2023.

\bibitem[Bickford~Smith et~al.(2023)Bickford~Smith, Kirsch, Farquhar, Gal, Foster, and Rainforth]{smith2023prediction}
Freddie Bickford~Smith, Andreas Kirsch, Sebastian Farquhar, Yarin Gal, Adam Foster, and Tom Rainforth.
\newblock Prediction-oriented {B}ayesian active learning.
\newblock In \emph{International Conference on Artificial Intelligence and Statistics}, pp.\  7331--7348, 2023.

\bibitem[Bickford~Smith et~al.(2025)Bickford~Smith, Kossen, Trollope, van~der Wilk, Foster, and Rainforth]{smith2025rethinking}
Freddie Bickford~Smith, Jannik Kossen, Eleanor Trollope, Mark van~der Wilk, Adam Foster, and Tom Rainforth.
\newblock Rethinking aleatoric and epistemic uncertainty.
\newblock In \emph{International Conference on Machine Learning}, 2025.
\newblock URL \url{https://openreview.net/forum?id=CY9MlORQs5}.

\bibitem[Billingsley(2013)]{billingsley2013convergence}
Patrick Billingsley.
\newblock \emph{Convergence of probability measures}.
\newblock John Wiley \& Sons, 2013.

\bibitem[Blau et~al.(2022)Blau, Bonilla, Chades, and Dezfouli]{blau2022optimizing}
Tom Blau, Edwin~V Bonilla, Iadine Chades, and Amir Dezfouli.
\newblock Optimizing sequential experimental design with deep reinforcement learning.
\newblock In \emph{International Conference on Machine Learning}, 2022.

\bibitem[Brown et~al.(2020)Brown, Mann, Ryder, Subbiah, Kaplan, Dhariwal, Neelakantan, Shyam, Sastry, Askell, Agarwal, Herbert-Voss, Krueger, Henighan, Child, Ramesh, Ziegler, Wu, Winter, Hesse, Chen, Sigler, Litwin, Gray, Chess, Clark, Berner, McCandlish, Radford, Sutskever, and Amodei]{brown2020language}
Brown, Mann, Ryder, Subbiah, Kaplan, Dhariwal, Neelakantan, Shyam, Sastry, Askell, Agarwal, Herbert-Voss, Krueger, Henighan, Child, Ramesh, Ziegler, Wu, Winter, Hesse, Chen, Sigler, Litwin, Gray, Chess, Clark, Berner, McCandlish, Radford, Sutskever, and Amodei.
\newblock Language models are few-shot learners.
\newblock In \emph{Advances in Neural Information Processing Systems}, 2020.

\bibitem[Cavagnaro et~al.(2010)Cavagnaro, Myung, Pitt, and Kujala]{cavagnaro2010adaptive}
Daniel~R Cavagnaro, Jay~I Myung, Mark~A Pitt, and Janne~V Kujala.
\newblock Adaptive design optimization: A mutual information-based approach to model discrimination in cognitive science.
\newblock \emph{Neural computation}, 22\penalty0 (4):\penalty0 887--905, 2010.

\bibitem[Chakraborty et~al.(2024)Chakraborty, Qiu, Yuan, Koppel, Huang, Manocha, Bedi, and Wang]{chakraborty2024maxminrlhfalignmentdiversehuman}
Souradip Chakraborty, Jiahao Qiu, Hui Yuan, Alec Koppel, Furong Huang, Dinesh Manocha, Amrit~Singh Bedi, and Mengdi Wang.
\newblock Maxmin-rlhf: Alignment with diverse human preferences.
\newblock In \emph{International Conference on Machine Learning}, 2024.

\bibitem[Chaloner \& Verdinelli(1995)Chaloner and Verdinelli]{chaloner1995bayesian}
Kathryn Chaloner and Isabella Verdinelli.
\newblock Bayesian experimental design: A review.
\newblock \emph{Statistical Science}, pp.\  273--304, 1995.

\bibitem[Chan et~al.(2025)Chan, Ge, Dobriban, Hassani, and Vidal]{chan2025conformal}
Kwan Ho~Ryan Chan, Yuyan Ge, Edgar Dobriban, Hamed Hassani, and Ren{\'e} Vidal.
\newblock Conformal information pursuit for interactively guiding large language models.
\newblock \emph{arXiv preprint arXiv:2507.03279}, 2025.

\bibitem[Chi et~al.(2024)Chi, Lin, Lin, and Klein]{chi2024clarinetaugmentinglanguagemodels}
Yizhou Chi, Jessy Lin, Kevin Lin, and Dan Klein.
\newblock Clarinet: Augmenting language models to ask clarification questions for retrieval.
\newblock \emph{arXiv preprint arXiv:2405.15784}, 2024.

\bibitem[Cooper et~al.(2025)Cooper, Wadhawan, Giorgi, Tan, and Liang]{cooper2025curious}
Michael Cooper, Rohan Wadhawan, John~Michael Giorgi, Chenhao Tan, and Davis Liang.
\newblock The curious language model: Strategic test-time information acquisition.
\newblock In \emph{Second Workshop on Test-Time Adaptation: Putting Updates to the Test! at ICML 2025}, 2025.
\newblock URL \url{https://openreview.net/forum?id=1Bfo9L5ayn}.

\bibitem[Desai \& Durrett(2020)Desai and Durrett]{desai-durrett-2020-calibration}
Shrey Desai and Greg Durrett.
\newblock Calibration of pre-trained transformers.
\newblock In Bonnie Webber, Trevor Cohn, Yulan He, and Yang Liu (eds.), \emph{Conference on Empirical Methods in Natural Language Processing}, pp.\  295--302, Online, November 2020. Association for Computational Linguistics.
\newblock \doi{10.18653/v1/2020.emnlp-main.21}.
\newblock URL \url{https://aclanthology.org/2020.emnlp-main.21/}.

\bibitem[Falck et~al.(2024)Falck, Wang, and Holmes]{falck2024incontextlearninglargelanguage}
Fabian Falck, Ziyu Wang, and Chris Holmes.
\newblock Is in-context learning in large language models {B}ayesian? {A} martingale perspective.
\newblock In \emph{International Conference on Machine Learning}, 2024.

\bibitem[Fong et~al.(2023)Fong, Holmes, and Walker]{fong2023martingale}
Edwin Fong, Chris Holmes, and Stephen~G Walker.
\newblock Martingale posterior distributions.
\newblock \emph{Journal of the Royal Statistical Society Series B: Statistical Methodology}, 85\penalty0 (5):\penalty0 1357--1391, 2023.

\bibitem[Foster(2021)]{foster2021variational}
Adam Foster.
\newblock \emph{Variational, {M}onte {C}arlo and policy-based approaches to {B}ayesian experimental design}.
\newblock PhD thesis, University of Oxford, 2021.

\bibitem[Foster et~al.(2021)Foster, Ivanova, Malik, and Rainforth]{foster2021deep}
Adam Foster, Desi~R Ivanova, Ilyas Malik, and Tom Rainforth.
\newblock Deep adaptive design: Amortizing sequential {B}ayesian experimental design.
\newblock In \emph{International Conference on Machine Learning}, pp.\  3384--3395, 2021.

\bibitem[Gal et~al.(2017)Gal, Islam, and Ghahramani]{gal2017deep}
Yarin Gal, Riashat Islam, and Zoubin Ghahramani.
\newblock Deep {B}ayesian active learning with image data.
\newblock In \emph{International Conference on Machine Learning}, pp.\  1183--1192, 2017.

\bibitem[Handa et~al.(2024)Handa, Gal, Pavlick, Goodman, Andreas, Tamkin, and Li]{handa2024bayesianpreferenceelicitationlanguage}
Kunal Handa, Yarin Gal, Ellie Pavlick, Noah Goodman, Jacob Andreas, Alex Tamkin, and Belinda~Z. Li.
\newblock Bayesian preference elicitation with language models.
\newblock \emph{arXiv preprint arXiv:2403.05534}, 2024.

\bibitem[Harper \& Konstan(2015)Harper and Konstan]{harper2015movielens}
F~Maxwell Harper and Joseph~A Konstan.
\newblock The movielens datasets: History and context.
\newblock \emph{Acm Transactions on Interactive Intelligent Systems}, 5\penalty0 (4):\penalty0 1--19, 2015.

\bibitem[Hedman et~al.(2025)Hedman, Ivanova, Guan, and Rainforth]{hedman2025stepdad}
Marcel Hedman, Desi~R. Ivanova, Cong Guan, and Tom Rainforth.
\newblock Step-{DAD}: Semi-amortized policy-based {B}ayesian experimental design.
\newblock In \emph{International Conference on Machine Learning}, 2025.
\newblock URL \url{https://openreview.net/forum?id=JRg8P2bX8P}.

\bibitem[Hirosawa et~al.(2024)Hirosawa, Harada, Mizuta, Sakamoto, Tokumasu, and Shimizu]{info:doi/10.2196/59267}
Takanobu Hirosawa, Yukinori Harada, Kazuya Mizuta, Tetsu Sakamoto, Kazuki Tokumasu, and Taro Shimizu.
\newblock Evaluating chatgpt-4's accuracy in identifying final diagnoses within differential diagnoses compared with those of physicians: Experimental study for diagnostic cases.
\newblock \emph{JMIR Form Res}, 8:\penalty0 e59267, Jun 2024.
\newblock ISSN 2561-326X.
\newblock \doi{10.2196/59267}.
\newblock URL \url{https://formative.jmir.org/2024/1/e59267}.

\bibitem[Hu et~al.(2024)Hu, Liu, Feng, Zhao, Ng, Luu, He, Koh, and Hooi]{hu2024uncertaintythoughtsuncertaintyawareplanning}
Zhiyuan Hu, Chumin Liu, Xidong Feng, Yilun Zhao, See-Kiong Ng, Anh~Tuan Luu, Junxian He, Pang~Wei Koh, and Bryan Hooi.
\newblock Uncertainty of thoughts: Uncertainty-aware planning enhances information seeking in large language models.
\newblock In \emph{Advances in Neural Information Processing Systems}, 2024.

\bibitem[Huan \& Marzouk(2016)Huan and Marzouk]{huan2016sequential}
Xun Huan and Youssef~M Marzouk.
\newblock Sequential {B}ayesian optimal experimental design via approximate dynamic programming.
\newblock \emph{arXiv preprint arXiv:1604.08320}, 2016.

\bibitem[Ivanova et~al.(2021)Ivanova, Foster, Kleinegesse, Gutmann, and Rainforth]{ivanova2021implicit}
Desi~R Ivanova, Adam Foster, Steven Kleinegesse, Michael~U Gutmann, and Tom Rainforth.
\newblock Implicit deep adaptive design: policy-based experimental design without likelihoods.
\newblock In \emph{Advances in Neural Information Processing Systems}, 2021.

\bibitem[Jacobsen et~al.(2025)Jacobsen, Cox, Griggio, and van Berkel]{Jacobsen_2025}
Rune~Møberg Jacobsen, Samuel~Rhys Cox, Carla~F. Griggio, and Niels van Berkel.
\newblock Chatbots for data collection in surveys: A comparison of four theory-based interview probes.
\newblock In \emph{CHI Conference on Human Factors in Computing Systems}, CHI ’25, pp.\  1–21. ACM, April 2025.
\newblock \doi{10.1145/3706598.3714128}.
\newblock URL \url{http://dx.doi.org/10.1145/3706598.3714128}.

\bibitem[Jha et~al.(2025)Jha, Arora, Watanabe, Yanagawa, Chen, Clark, Bhavya, Verma, Kumar, Kitahara, Zheutlin, Takano, Pathak, George, Wu, Turkkan, Vanloo, Nidd, Dai, Chatterjee, Gupta, Samanta, Aggarwal, Lee, Murali, wook Ahn, Kar, Rahane, Fonseca, Paradkar, Deng, Moogi, Mohapatra, Abe, Narayanaswami, Xu, Varshney, Mahindru, Sailer, Shwartz, Sow, Fuller, and Puri]{jha2025itbenchevaluatingaiagents}
Saurabh Jha, Rohan Arora, Yuji Watanabe, Takumi Yanagawa, Yinfang Chen, Jackson Clark, Bhavya Bhavya, Mudit Verma, Harshit Kumar, Hirokuni Kitahara, Noah Zheutlin, Saki Takano, Divya Pathak, Felix George, Xinbo Wu, Bekir~O. Turkkan, Gerard Vanloo, Michael Nidd, Ting Dai, Oishik Chatterjee, Pranjal Gupta, Suranjana Samanta, Pooja Aggarwal, Rong Lee, Pavankumar Murali, Jae wook Ahn, Debanjana Kar, Ameet Rahane, Carlos Fonseca, Amit Paradkar, Yu~Deng, Pratibha Moogi, Prateeti Mohapatra, Naoki Abe, Chandrasekhar Narayanaswami, Tianyin Xu, Lav~R. Varshney, Ruchi Mahindru, Anca Sailer, Laura Shwartz, Daby Sow, Nicholas C.~M. Fuller, and Ruchir Puri.
\newblock Itbench: Evaluating ai agents across diverse real-world it automation tasks.
\newblock In \emph{International Conference on Machine Learning}, 2025.

\bibitem[Kadavath et~al.(2022)Kadavath, Conerly, Askell, Henighan, Drain, Perez, Schiefer, Hatfield-Dodds, DasSarma, Tran-Johnson, Johnston, El-Showk, Jones, Elhage, Hume, Chen, Bai, Bowman, Fort, Ganguli, Hernandez, Jacobson, Kernion, Kravec, Lovitt, Ndousse, Olsson, Ringer, Amodei, Brown, Clark, Joseph, Mann, McCandlish, Olah, and Kaplan]{kadavath2022languagemodelsmostlyknow}
Saurav Kadavath, Tom Conerly, Amanda Askell, Tom Henighan, Dawn Drain, Ethan Perez, Nicholas Schiefer, Zac Hatfield-Dodds, Nova DasSarma, Eli Tran-Johnson, Scott Johnston, Sheer El-Showk, Andy Jones, Nelson Elhage, Tristan Hume, Anna Chen, Yuntao Bai, Sam Bowman, Stanislav Fort, Deep Ganguli, Danny Hernandez, Josh Jacobson, Jackson Kernion, Shauna Kravec, Liane Lovitt, Kamal Ndousse, Catherine Olsson, Sam Ringer, Dario Amodei, Tom Brown, Jack Clark, Nicholas Joseph, Ben Mann, Sam McCandlish, Chris Olah, and Jared Kaplan.
\newblock Language models (mostly) know what they know.
\newblock \emph{arXiv preprint arXiv:2207.05221}, 2022.

\bibitem[Kestin et~al.(2025)Kestin, Miller, Klales, Milbourne, and Ponti]{kestin2025ai}
Greg Kestin, Kelly Miller, Anna Klales, Timothy Milbourne, and Gregorio Ponti.
\newblock Ai tutoring outperforms in-class active learning: an rct introducing a novel research-based design in an authentic educational setting.
\newblock \emph{Scientific Reports}, 15\penalty0 (1):\penalty0 17458, 2025.

\bibitem[Kobalczyk et~al.(2025)Kobalczyk, Astorga, Liu, and van~der Schaar]{kobalczyk2025active}
Katarzyna Kobalczyk, Nicolas Astorga, Tennison Liu, and Mihaela van~der Schaar.
\newblock Active task disambiguation with llms.
\newblock In \emph{International Conference on Learning Representations}, 2025.

\bibitem[Kossen et~al.(2024)Kossen, Gal, and Rainforth]{kossen2024incontext}
Jannik Kossen, Yarin Gal, and Tom Rainforth.
\newblock In-context learning learns label relationships but is not conventional learning.
\newblock In \emph{The Twelfth International Conference on Learning Representations}, 2024.
\newblock URL \url{https://openreview.net/forum?id=YPIA7bgd5y}.

\bibitem[Laban et~al.(2025)Laban, Hayashi, Zhou, and Neville]{laban2025llmslostmultiturnconversation}
Philippe Laban, Hiroaki Hayashi, Yingbo Zhou, and Jennifer Neville.
\newblock Llms get lost in multi-turn conversation.
\newblock \emph{arXiv preprint arXiv:2505.06120}, 2025.

\bibitem[Lee et~al.(2024)Lee, Peng, Goldberg, Rosenthal, Kotcher, Maibach, and Leiserowitz]{lee2024can}
Sanguk Lee, Tai-Quan Peng, Matthew~H Goldberg, Seth~A Rosenthal, John~E Kotcher, Edward~W Maibach, and Anthony Leiserowitz.
\newblock Can large language models estimate public opinion about global warming? an empirical assessment of algorithmic fidelity and bias.
\newblock \emph{PLoS Climate}, 3\penalty0 (8):\penalty0 e0000429, 2024.

\bibitem[Li et~al.(2025{\natexlab{a}})Li, Kim, and Wang]{li2025questbench}
Belinda~Z Li, Been Kim, and Zi~Wang.
\newblock Questbench: Can llms ask the right question to acquire information in reasoning tasks?
\newblock \emph{arXiv preprint arXiv:2503.22674}, 2025{\natexlab{a}}.

\bibitem[Li et~al.(2025{\natexlab{b}})Li, Tamkin, Goodman, and Andreas]{li2025eliciting}
Belinda~Z. Li, Alex Tamkin, Noah Goodman, and Jacob Andreas.
\newblock Eliciting human preferences with language models.
\newblock In \emph{The Thirteenth International Conference on Learning Representations}, 2025{\natexlab{b}}.
\newblock URL \url{https://openreview.net/forum?id=LvDwwAgMEW}.

\bibitem[Li et~al.(2025{\natexlab{c}})Li, Shen, Yao, Ding, Miao, Krishnan, and Padman]{li2025singleturnsurveymultiturninteractions}
Yubo Li, Xiaobin Shen, Xinyu Yao, Xueying Ding, Yidi Miao, Ramayya Krishnan, and Rema Padman.
\newblock Beyond single-turn: A survey on multi-turn interactions with large language models.
\newblock \emph{arXiv preprint arXiv:2504.04717}, 2025{\natexlab{c}}.

\bibitem[Lindley(1972)]{lindley1972bayesian}
Lindley.
\newblock \emph{{Bayesian} Statistics: a Review}.
\newblock Society for Industrial and Applied Mathematics, 1972.

\bibitem[Lindley(1956)]{lindley1956measure}
Dennis~V Lindley.
\newblock On a measure of the information provided by an experiment.
\newblock \emph{The Annals of Mathematical Statistics}, 27\penalty0 (4):\penalty0 986--1005, 1956.

\bibitem[Liu et~al.(2024{\natexlab{a}})Liu, Huang, Xiao, Sha, Wu, Liu, Wang, and Chen]{liu2024socraticlm}
Jiayu Liu, Zhenya Huang, Tong Xiao, Jing Sha, Jinze Wu, Qi~Liu, Shijin Wang, and Enhong Chen.
\newblock Socraticlm: Exploring socratic personalized teaching with large language models.
\newblock \emph{Advances in Neural Information Processing Systems}, 37:\penalty0 85693--85721, 2024{\natexlab{a}}.

\bibitem[Liu et~al.(2024{\natexlab{b}})Liu, Lin, Hewitt, Paranjape, Bevilacqua, Petroni, and Liang]{liu2023lost}
Nelson~F Liu, Kevin Lin, John Hewitt, Ashwin Paranjape, Michele Bevilacqua, Fabio Petroni, and Percy Liang.
\newblock Lost in the middle: How language models use long contexts.
\newblock \emph{Transactions of the Association for Computational Linguistics}, 12, 2024{\natexlab{b}}.

\bibitem[Lu et~al.(2024)Lu, Lu, Lange, Foerster, Clune, and Ha]{lu2024aiscientistfullyautomated}
Chris Lu, Cong Lu, Robert~Tjarko Lange, Jakob Foerster, Jeff Clune, and David Ha.
\newblock The ai scientist: Towards fully automated open-ended scientific discovery.
\newblock \emph{arXiv preprint arXiv:2408.06292}, 2024.

\bibitem[MacKay(1992)]{mackay1992information}
David~JC MacKay.
\newblock Information-based objective functions for active data selection.
\newblock \emph{Neural computation}, 4\penalty0 (4):\penalty0 590--604, 1992.

\bibitem[Mandal et~al.(2025)Mandal, Soni, Zaki, Smedskjaer, Wondraczek, Wondraczek, Gosvami, and Krishnan]{mandal2025autonomousmicroscopyexperimentslarge}
Indrajeet Mandal, Jitendra Soni, Mohd Zaki, Morten~M. Smedskjaer, Katrin Wondraczek, Lothar Wondraczek, Nitya~Nand Gosvami, and N.~M.~Anoop Krishnan.
\newblock Autonomous microscopy experiments through large language model agents.
\newblock \emph{arXiv preprint: arXiv2501.10385}, 2025.

\bibitem[Mazzaccara et~al.(2024)Mazzaccara, Testoni, and Bernardi]{mazzaccara2024learningaskinformativequestions}
Davide Mazzaccara, Alberto Testoni, and Raffaella Bernardi.
\newblock Learning to ask informative questions: Enhancing llms with preference optimization and expected information gain.
\newblock In \emph{Findings of the Association for Computational Linguistics: EMNLP}, 2024.

\bibitem[Min et~al.(2022)Min, Lyu, Holtzman, Artetxe, Lewis, Hajishirzi, and Zettlemoyer]{min2022rethinking}
Sewon Min, Xinxi Lyu, Ari Holtzman, Mikel Artetxe, Mike Lewis, Hannaneh Hajishirzi, and Luke Zettlemoyer.
\newblock Rethinking the role of demonstrations: What makes in-context learning work?
\newblock \emph{arXiv preprint arXiv:2202.12837}, 2022.

\bibitem[Neiswanger et~al.(2021)Neiswanger, Wang, and Ermon]{neiswanger2021bayesian}
Willie Neiswanger, Ke~Alexander Wang, and Stefano Ermon.
\newblock Bayesian algorithm execution: Estimating computable properties of black-box functions using mutual information.
\newblock In \emph{International Conference on Machine Learning}, pp.\  8005--8015, 2021.

\bibitem[OpenAI(2024)]{openai2024gpt4ocard}
OpenAI.
\newblock Gpt-4o system card.
\newblock \emph{arXiv preprint arXiv:2410.21276}, 2024.

\bibitem[{OpenAI}(2025)]{openai_o3_2025}
{OpenAI}.
\newblock {OpenAI} o3 and o4-mini system card.
\newblock System card, OpenAI, April 2025.
\newblock URL \url{https://cdn.openai.com/pdf/2221c875-02dc-4789-800b-e7758f3722c1/o3-and-o4-mini-system-card.pdf}.
\newblock Accessed 2025-07-15.

\bibitem[Ouyang et~al.(2022)Ouyang, Wu, Jiang, Almeida, Wainwright, Mishkin, Zhang, Agarwal, Slama, Ray, et~al.]{ouyang2022training}
Long Ouyang, Jeffrey Wu, Xu~Jiang, Diogo Almeida, Carroll Wainwright, Pamela Mishkin, Chong Zhang, Sandhini Agarwal, Katarina Slama, Alex Ray, et~al.
\newblock Training language models to follow instructions with human feedback.
\newblock \emph{Advances in Neural Information Processing Systems}, 35:\penalty0 27730--27744, 2022.

\bibitem[Paninski(2003)]{paninski2003estimation}
Liam Paninski.
\newblock Estimation of entropy and mutual information.
\newblock \emph{Neural computation}, 15\penalty0 (6):\penalty0 1191--1253, 2003.

\bibitem[Patil et~al.(2025)Patil, Mao, Yan, Ji, Suresh, Stoica, and Gonzalez]{patil2025the}
Shishir~G Patil, Huanzhi Mao, Fanjia Yan, Charlie Cheng-Jie Ji, Vishnu Suresh, Ion Stoica, and Joseph~E. Gonzalez.
\newblock The berkeley function calling leaderboard ({BFCL}): From tool use to agentic evaluation of large language models.
\newblock In \emph{International Conference on Machine Learning}, 2025.
\newblock URL \url{https://openreview.net/forum?id=2GmDdhBdDk}.

\bibitem[Piriyakulkij et~al.(2023)Piriyakulkij, Kuleshov, and Ellis]{piriyakulkij2023active}
Wasu~Top Piriyakulkij, Volodymyr Kuleshov, and Kevin Ellis.
\newblock Active preference inference using language models and probabilistic reasoning.
\newblock \emph{arXiv preprint arXiv:2312.12009}, 2023.

\bibitem[Rainforth(2017)]{rainforth2017automating}
Tom Rainforth.
\newblock \emph{Automating inference, learning, and design using probabilistic programming}.
\newblock PhD thesis, University of Oxford, 2017.

\bibitem[Rainforth et~al.(2024)Rainforth, Foster, Ivanova, and Bickford~Smith]{rainforth2024modern}
Tom Rainforth, Adam Foster, Desi~R Ivanova, and Freddie Bickford~Smith.
\newblock Modern {B}ayesian experimental design.
\newblock \emph{Statistical Science}, 39\penalty0 (1):\penalty0 100--114, 2024.

\bibitem[Rao et~al.(1945)]{rao1945information}
C~Radhakrishna Rao et~al.
\newblock Information and the accuracy attainable in the estimation of statistical parameters.
\newblock \emph{Bull. Calcutta Math. Soc}, 37\penalty0 (3):\penalty0 81--91, 1945.

\bibitem[Savage(1954)]{savage1954foundations}
Leonard~J Savage.
\newblock \emph{The foundations of statistics.}
\newblock John Wiley \& Sons, 1954.

\bibitem[Sebastiani \& Wynn(2000)Sebastiani and Wynn]{sebastiani2000maximum}
Paola Sebastiani and Henry~P Wynn.
\newblock Maximum entropy sampling and optimal {B}ayesian experimental design.
\newblock \emph{Journal of the Royal Statistical Society: Series B (Statistical Methodology)}, 62\penalty0 (1):\penalty0 145--157, 2000.

\bibitem[Trivedi et~al.(2024)Trivedi, Gulati, Molenschot, Rajeev, Ramamurthy, Stevens, Chaudhery, Jambholkar, Zou, and Rajani]{trivedi2024selfrationalizationimprovesllmfinegrained}
Prapti Trivedi, Aditya Gulati, Oliver Molenschot, Meghana~Arakkal Rajeev, Rajkumar Ramamurthy, Keith Stevens, Tanveesh~Singh Chaudhery, Jahnavi Jambholkar, James Zou, and Nazneen Rajani.
\newblock Self-rationalization improves llm as a fine-grained judge.
\newblock \emph{arXiv preprint arXiv:2410.05495}, 2024.

\bibitem[Wang et~al.(2025)Wang, Zollo, Zemel, and Namkoong]{wang2025adaptiveelicitationlatentinformation}
Jimmy Wang, Thomas Zollo, Richard Zemel, and Hongseok Namkoong.
\newblock Adaptive elicitation of latent information using natural language.
\newblock \emph{arXiv preprint arXiv:2504.04204}, 2025.

\bibitem[Wu et~al.(2025)Wu, Galley, Peng, Cheng, Li, Dou, Cai, Zou, Leskovec, and Gao]{wu2025collabllmpassiverespondersactive}
Shirley Wu, Michel Galley, Baolin Peng, Hao Cheng, Gavin Li, Yao Dou, Weixin Cai, James Zou, Jure Leskovec, and Jianfeng Gao.
\newblock Collabllm: From passive responders to active collaborators.
\newblock In \emph{International Conference on Machine Learning}, 2025.

\bibitem[Yao et~al.(2023{\natexlab{a}})Yao, Yu, Zhao, Shafran, Griffiths, Cao, and Narasimhan]{yao2023tree}
Shunyu Yao, Dian Yu, Jeffrey Zhao, Izhak Shafran, Tom Griffiths, Yuan Cao, and Karthik Narasimhan.
\newblock Tree of thoughts: Deliberate problem solving with large language models.
\newblock In \emph{Advances in Neural Information Processing Systems}, 2023{\natexlab{a}}.

\bibitem[Yao et~al.(2023{\natexlab{b}})Yao, Zhao, Yu, Du, Shafran, Narasimhan, and Cao]{yao2023react}
Shunyu Yao, Jeffrey Zhao, Dian Yu, Nan Du, Izhak Shafran, Karthik Narasimhan, and Yuan Cao.
\newblock React: Synergizing reasoning and acting in language models, 2023{\natexlab{b}}.
\newblock URL \url{https://arxiv.org/abs/2210.03629}.

\bibitem[Zhang et~al.(2024)Zhang, Lu, and Jaitly]{zhang2024probingmultiturnplanningcapabilities}
Yizhe Zhang, Jiarui Lu, and Navdeep Jaitly.
\newblock Probing the multi-turn planning capabilities of llms via 20 question games.
\newblock In \emph{Proceedings of the Association for Computational Linguistics}, 2024.

\bibitem[Zhang et~al.(2025)Zhang, Li, Cui, Cai, Liu, Fu, Huang, Zhao, Zhang, Chen, et~al.]{zhang2023siren}
Yue Zhang, Yafu Li, Leyang Cui, Deng Cai, Lemao Liu, Tingchen Fu, Xinting Huang, Enbo Zhao, Yu~Zhang, Yulong Chen, et~al.
\newblock Siren's song in the ai ocean: a survey on hallucination in large language models.
\newblock \emph{ACM Transactions on Information Systems}, 42\penalty0 (2), 2025.

\bibitem[Zhu et~al.(2025)Zhu, Wang, and Wang]{zhu2025judgelmfinetunedlargelanguage}
Lianghui Zhu, Xinggang Wang, and Xinlong Wang.
\newblock Judgelm: Fine-tuned large language models are scalable judges.
\newblock In \emph{International Conference on Learning Representations}, 2025.

\end{thebibliography}
\newpage
\appendix

\startcontents[appendix]
\printcontents[appendix]{}{1}{\section*{Appendix contents}}

\clearpage
\section{Discussion relating to \Cref{sec:method}}  %

\subsection{Updating the likelihood}
\label{app:updating_likelihood}

The success of BED-LLM hinges on our ability to update our joint distribution. 
As mentioned in \textsection\ref{sec:method}, we choose not to 
update the likelihood model as more data is gathered, that is, our likelihood in the sequential setting will be $\seqlikellm$ instead of $\pllm(\data_t;[\history,\theta,\design_t])$.
The main rationale of this choice is that for many problems our beliefs on $\theta$ capture all the required information to predict $\data|\design$, hence including the history is adding unnecessary context that could influence the LLM's behavior in undesirable ways. See \Cref{app:updating_likelihood} for a results comparison of static and updated likelihoods on the 20 Questions game.
However, it is important to note that $\pllm(\data_t;[\history,\theta,\design_t])$ should be used instead for problems where $\theta$ will not capture all information from previous data, e.g.~if $\theta$ is a binary value corresponding to whether we reject a null hypothesis, or is the answer to a particular other question of interest.

\subsection{Estimating EIG for each question}

\begin{figure}[h]
    \centering
    {\small
    
    \begin{minipage}{0.98\linewidth}
    
      \begin{minipage}{.495\textwidth}
        \begin{qcardA}{Question 1}
          \textbf{Question:}\\
          Which ice~cream flavor feels like the best match for this user?
    
          \vspace{0.45em}
          \textbf{Choose one option:}
          \begin{enumerate}[label=\textbf{\Alph*.}, leftmargin=*, itemsep=1pt, topsep=1pt]
            \item Vanilla
            \item Dark Chocolate
            \item Strawberry Swirl
            \item Mint~Chocolate~Chip
          \end{enumerate}
    
          \vspace{0.2em}
          \begin{metricA}
            \centering
             \textbf{Predictive Entropy: Very High}\\[-1pt]
             \textbf{EIG: 0}
          \end{metricA}
        \end{qcardA}
      \end{minipage}\hfill
      \begin{minipage}{.495\textwidth}
        \begin{qcardB}{Question 2}
          \textbf{Question:}\\
          Which film genre does the user most prefer?\\
    
          \vspace{0.45em}
          \textbf{Choose one option:}
          \begin{enumerate}[label=\textbf{\Alph*.}, leftmargin=*, itemsep=1pt, topsep=1pt]
            \item Action
            \item Sci-Fi
            \item Comedy
            \item Horror
          \end{enumerate}
    
          \vspace{0.2em}
          \begin{metricB}
            \centering
             \textbf{Predictive Entropy: High }\\[-1pt]
             \textbf{EIG: High}
          \end{metricB}
        \end{qcardB}
      \end{minipage}
    \end{minipage}
    }
    \caption{Predictive entropy vs.\ expected information gain (EIG) in a film‐preferences elicitation task. Left: very high predictive entropy (answer is completely unknown) but EIG $= 0$ because the answer provides no insight into the user's film preferences. %
    Right: both predictive entropy and EIG are high as the answer is uncertain, but different answers would lead to markedly different posterior updates, 
    making it informative for learning film preferences.
    This thus demonstrates how the two criteria can select different questions.
    }
    \label{fig:questions_side}
\end{figure}

\subsubsection{Predictive entropy is not a good approximation of EIG}
\label{app:entropy}

As discussed in \textsection\ref{sec:est}, previous information-based query selection mechanisms have assumed that responses are deterministic given $\theta$ and $\design$. This implies that the expected entropy of the likelihood, $\E_{p(\theta;\history)}[\mathrm{H}[p(\data_{t+1}|\theta;\design_{t+1},\history)]]$, is constant over designs, meaning that maximizing EIG is equivalent to maximizing the marginal predictive entropy,  $\mathrm{H}[\E_{p(\theta;\history)}[p(\data_{t}|\theta;\design_{t},\history)]]$.

In practice, the expected likelihood entropy can and will vary across designs. This variability in the expected likelihood entropy can be crucial in selecting good designs.
A concrete example helps highlight how predictive entropy can differ significantly from EIG.
\Cref{fig:questions_side} shows two candidate questions that could be asked to elicit film preference. Question 1 has high predictive entropy: in a randomly selected group of people, we would expect high variation in ice cream preference (regardless of the individual's film preferences). However, since ice cream preference is unrelated to film preference, the answer would not help us narrow down our hypothesis space, and the EIG is zero.

This is also supported by evidence in
our experiments (\textsection\ref{sec:experiments}). Both the Split baseline, and the Entropy ablation, assume a deterministic likelihood; in particular, the Entropy ablation uses the same estimator of the predictive entropy as \name. In both cases, we see the performance significantly degrades relative to using the full EIG.  Further, omitting the expected likelihood entropy term provides no meaningful computational saving---the same LLM evaluations are used for the top and bottom lines of~\Cref{eq:est-rb}, hence doing the full estimate of the EIG requires no additional LLM calls to be made.

\subsubsection{EIG estimator}

One might 
be tempted to replace $\hat{p}(\data_{t+1} ; [\history,\design_{t+1}])$ with $\pllm(\data_{t+1};[\history,\design_{t+1}])$ in the EIG estimation in \Cref{eq:est-rb}, 
as the two essentially offer alternative predictive distributions for the outcome.
We also advise against this, though, noting that it again provides no meaningful computational benefits (unless one also assumes a deterministic likelihood, but this would then mean we no longer consider $\theta$ at all).
A key reason for avoiding this substitution is that it would mean we are no longer estimating a true EIG: the inconsistency between the likelihood and the marginal data distribution means there is no longer a joint model where we are minimising our expected uncertainty in $\theta$.
We also find that the LLM process of sampling $\theta$ from $\thetafilter$ followed by $\data$ from $\seqlikellm$ tends to give a better uncertainty over responses than sampling $\data$ directly from $\pllm(\data_{t+1};[\history,\design_{t+1}])$.

\subsection{Prior construction and belief updating}
\label{app:sBED}

In \Cref{sec:sBED}, we argued that \baseline in-context updating is not sufficient for updating our beliefs: We fail to fully incorporate the information from the history $h_t$, and we often have overconfident distributions. The shortfalls of in-context learning in such settings have also previously be noted by, for example,
\citep{liu2023lost, zhang2023siren, zhang2024probingmultiturnplanningcapabilities}.
We posit two reasons why they likely struggle in such settings.
First, the information from the different examples in the history are generally highly distinct in these information-gathering settings (indeed, this is part of our aim in adaptively design informative questions), making it harder for the LLM to appropriately reconcile all the provided information than in many other uses of in--context learning.
Second, $\theta$ will often represent a user--specific variable that cannot easily be predicted from any data other than the user's responses to questions: it has been argued that much of the success of in--context learning in LLMs is down to improving problem specification and linking the requested task to data it has seen in its training, rather than truly ``learning'' from the provided examples~\citep{min2022rethinking,kossen2024incontext}, but the history in our setting is rarely helpful for this due to its user--specific nature.

\section{Discussion relating to \Cref{sec:justification}}
\label{app:faithfulness_cont}

\subsection{An alternative view on the faithfulness of conditional distributions}

Another way of viewing the distinction between the prior--likelihood and data--estimation constructions is in which of the EIG forms,~\Cref{eq:bayes_eig_theta} or~\Cref{eq:bald}, we center our reasoning.
For a given joint model, the two are, of course, mathematically equivalent.
However, they give us different ways of thinking about what it means to maximize the EIG: reducing entropy in $\theta$ from seeing $\data$, or reducing entropy in $\data$ from seeing $\theta$.
This, in turn, gives us a way to reason about how appropriate our joint model is.
When we choose to use one of $\likellm$ or $\postllm$, we are centering our reasoning around the entropy of this quantity making sense, while allowing the other entropy in the other form to be implicitly defined from the resulting joint distribution; because the two forms are equivalent, we know that if our explicit form is suitable/unsuitable, the implicit form will be as well.
If, for example, we directly fix the form of $\postllm$ using our LLM's predictive distribution, we are also directly relying on its expected entropy being a meaningful measure of design quality.
If $\theta$ is high-dimensional and predominantly free-form, the resulting entropy produced by the LLM is unlikely to be meaningful and using the data--estimation pairing is unlikely to produce an effective strategy.
However, if $\data$ is instead quite constrained, the LLM can produce a meaningful entropy over it, and choosing a model based on the prior--likelihood pairing is likely to \emph{implicitly define} a meaningful distribution, and thus entropy, on $\theta$.
Conversely, if $\theta$ is constrained and $\data$ is free form, the opposite will hold instead.

\subsection{Choice of $\theta$} 
An important corollary of this reasoning is that it can be important to be careful in our choice of exactly what we take $\theta$ to be, especially if we are using the data--estimation formulation.  
In particular, it is essential for entropy in the space of $\theta$ to form a meaningful notion of uncertainty, even if this entropy is not being measured through the LLM's predictive distribution of $\theta$ directly.
Thus, while $\theta$ inherently represents what we are trying to learn about and should always be set up as such, if there is flexibility in how exactly we formulate it, we should be careful to choose a form that yields an appropriate uncertainty measure.
For example, if the LLM is trying to clarify what code a user wishes it to generate, we could either choose $\theta$ to be the code itself or, following~\citep{neiswanger2021bayesian,smith2023prediction}, the output the code produces.
Here the entropy over code outputs induced by our distribution on code is likely to be a much better measure of uncertainty than the entropy of the raw code itself, given that there are multiple ways one can code the same operation.

\subsection{Alignment between EIG and belief updating procedure}\label{sec:updating_hypotheses_alignment}

If our ultimate goal is to minimize uncertainty in $\theta$, as measured by its entropy, then we can use the expected uncertainty reduction framework of \citet{smith2025rethinking} to provide insights into how well our EIG formulation and belief updating procedures align.
To simplify discussions, 
for now we consider the setting where we choose a single question $x$ and obtain a response $y$.
Following \citet{smith2025rethinking}, we can think of the ``true'' optimal design as selecting
\begin{align}
    \design^*_{\mathrm{true}} = \argmin_\design \E_{p_{\mathrm{true}}(\data;\design)}\left[\mathrm{H}[\updatedmodel]\right],\label{eq:fbs}
\end{align}
where $p_{\mathrm{true}}(\data;\design)$ is the true response distribution and $\updatedmodel$ is our belief state after the experiment. 

Note here that true optimal design has no direct dependency on our current beliefs about $\theta$; %
it only depends on $p_{\mathrm{true}}(\data;\design)$ and the hypothetical beliefs we produce for given observed data, $\updatedmodel$.
Thus, we can now see that our choice of joint model corresponds to different choices for approximating these quantities.
Assuming that the LLM distribution is used directly for the conditional as per~\Cref{sec:justification}, we thus have that
\begin{itemize}[labelindent=-5pt,leftmargin=!,topsep=0pt]
    \item the prior--likelihood pairing corresponds to using the approximations $p_{\mathrm{true}}(\data;\design) \approx \int p(\theta) \pllm(\data;[\theta,\design])d\theta$ and $\updatedmodel \approx p(\theta) \pllm(\data;[\theta,\design])/\int p(\theta) \pllm(\data;[\theta,\design]) d\theta$; and
    \item the data--estimation pairing corresponds to directly specifying a model for $p_{\mathrm{true}}(\data;\design)$ and then using the approximation $\updatedmodel \approx \pllm (\theta;[\design,\data])$.
\end{itemize}
The appropriateness of each of these options, therefore, comes down to how faithful these approximations are respectively to the true data distribution, $p_{\mathrm{true}}(\data;\design)$, and how we actually derive our belief distribution on $\theta$ in practice once we have seen the new data.

The former of these considerations is difficult to control for as we simply do not know the true response distribution and it is hard to say which approach will thus estimate it best (though we can refer to the discussion in \Cref{sec:justification} to determine which best matches our \textit{beliefs} about the true response distribution).
However, we do know upfront how we plan to derive our belief distribution on $\theta$ in practice, so we can use this to guide which joint model we formulate our EIG from.
Namely, we observe that
(a) using the prior--likelihood EIG pairing equates to assuming we will make a \emph{Bayesian update to our beliefs} on $\theta$ using the likelihood $\pllm(\data;[\theta,\design])$, and
(b) using the data--estimation EIG pairing equates to assuming we will make an \emph{in-context update to our beliefs} on $\theta$, as we are treating $\updatedmodel$ as $\pllm (\theta;[\design,\data])$.

Our preference between the pairings should therefore be guided in part by \emph{how we plan to update the model in practice}.
In particular, if we plan to make pure Bayesian updates, then the prior--likelihood formulation will tend to yield an EIG that is more faithful to our updating procedure, while if we only make simple in-context updates, the data--estimation formulation will tend to yield a more faithful EIG instead.

The update we use in practice, namely taking $\updatedmodel=p_f(\theta;[\design,\data])$ as outlined in~\Cref{sec:sBED}, can be seen as being somewhere between the in-context and Bayesian updating: we initially sample from $\pllm(\theta;[\design,\data])$, but then perform filtering and other steps. 
The relative extent to which it resembles each will be problem--dependent and again be linked to how much we trust the LLM to capture uncertainty in the space of $\theta$ vs.~$\data$.

For the settings we consider, we expect $p_f(\theta;[\design,\data])$ to generally be better approximated by a Bayesian update than an in-context update, aligning with our decision to use the prior--likelihood formulation.
The reasons for this are that a) the filtering often removes a large proportion of the generated samples, especially at later experiment turns, with $\pllm(\theta;\history)$ not fully incorporating information from the history; b) the maintaining of the set of one consistent hypotheses from one turn to the next encourages a more Bayesian behavior, with samples persisting unless contradicted by a new likelihood term; and c) the typical premature overconfidence of $\pllm(\theta;\history)$ to a small number of hypotheses means it is typically unrepresentative of our beliefs.

\looseness=-1
These theoretical benefits are perhaps secondary to the more practical benefits from the ease of constructing an appropriate model in the prior--likelihood formulation and avoiding direct uncertainty estimation in the space of $\theta$.
Nonetheless, they help confirm that our choices have not induced unnecessary mismatch between the EIG formulation and our updating procedure.

The picture here can get a somewhat more complicated once we move into the sequential BED setting.
Here, our ultimate aim is actually to minimize $\mathrm{H}[p(\theta;h_T)]$ at some final future horizon $T$.
Now, we only care about intermediary belief states $p(\theta;h_t)$ through their aid in future decision making toward the goal of minimizing the final entropy.
Thus, even if we are working with in-context updates, it might be the case that $p(\theta;h_t)$ only starts to produce a meaningful entropy once we have seen enough data to sufficiently narrow down the possibilities on $\theta$.
The optimal behavior in such settings would be to learn a policy that directly targets this final belief state instead of sequentially targeting the incremental EIGs.
However, this will typically not be computationally feasible in practice and we instead need to resort to a myopic decision-making strategy.
It might thus still be better to use the prior--likelihood formulation in such myopic decision making settings, even if we are sequentially updating our beliefs on $\theta$ through in-context updates, if this allows us to better guide the sequential decisions towards our final objective.
The coherence of Bayesian updating means that the converse is unlikely to be true, so this provides further evidence towards using the prior--likelihood formulation.

\section{Extended related work}
\label{app:extended_related}

\paragraph{Information-based question answering with LLMs}
Several recent works have (explicitly or implicitly) looked at information gathering with LLMs. Most of these can be framed in a BED setting, with a \emph{deterministic likelihood} \citep{piriyakulkij2023active,hu2024uncertaintythoughtsuncertaintyawareplanning,kobalczyk2025active,cooper2025curious}, and can be seen as variants of our Split baseline. 
\citet{piriyakulkij2023active} use a deterministic 0/1 answer likelihood $p(a | x, q)$ via the LLM to prune items from a pre-enumerated finite set given a candidate question $q$. The question is selected by minimizing expected posterior entropy. They model user preferences with a binary ground truth, which would not be applicable in preference-elicitation scenarios with nebulous user profiles. Similarly, ~\citet{hu2024uncertaintythoughtsuncertaintyawareplanning}
use a deterministic likelihood to minimize entropy over a finite set $\Omega$ in a closed-world setting. \citet{kobalczyk2025active} target ambiguous task specifications in open-ended generation tasks by sampling a set of hypotheses (placing a uniform prior over them) and viewing each question as a deterministic partition over those samples, looking for questions that split the samples roughly evenly. \citet{cooper2025curious} compute posterior entropy over a working set of top $k$ hypotheses (without filtering) through heuristic pruning. %

\citet{wang2025adaptiveelicitationlatentinformation} avoid the pitfall of deterministic likelihoods. They use a data--estimation framework to estimate EIG, focusing on scenarios where the target can be expressed as a predefined series of multiple-choice questions. Their approach relies on meta-training a predictive language model on historic question/answer pairs, and so is not directly comparable with BED-LLM which requires no additional training or data. \citet{chan2025conformal} do not model likelihoods or posterior beliefs, instead they rely on the expected size of conformal prediction sets as a surrogate uncertainty metric. This requires the use of an additional calibration dataset, and is confined to closed-world settings with a finite label set and pre-defined queries.

\paragraph{Post-training LLMs for improved information gathering}
Rather than augmenting a frozen LLM with the ability to estimate utility functions, some works have instead aimed to post-train an LLM to improve its ability to ask questions \citep{zhang2024probingmultiturnplanningcapabilities,wu2025collabllmpassiverespondersactive,andukuri2024star}. Most do not explicitly consider informativeness of questions: \citet{zhang2024probingmultiturnplanningcapabilities} and \citet{wu2025collabllmpassiverespondersactive} use reinforcement learning techniques to reward generations that quickly lead to the correct answer, and \citet{andukuri2024star} builds on \citet{li2025eliciting} by fine-tuning on successful traces. \citet{mazzaccara2024learningaskinformativequestions} do indirectly incorporate uncertainty, also using a deterministic likelihood: they use predictive entropy to identify informative questions, and then either fine-tune on the highest-entropy question, or perform DPO comparing the highest-entropy question with a lower-entropy question.  We do not address fine-tuning in this work, focusing instead on exploring the correct way to formulate BED using LLMs.

\paragraph{Combining LLMs with parametric models}
As discussed in~\Cref{sec:justification}, a key challenge in adapting BED to the LLM setting is in aligning the expected information gain with the actual uncertainties extracted from the LLM after updating. \citet{handa2024bayesianpreferenceelicitationlanguage} take a different approach to this problem by using the LLM to generate features for an external conventional Bayesian joint model 
(in their case, a linear Bradley--Terry model), rather than deriving their joint model more directly from the LLM itself. This can be a good choice when the problem is well-bounded and we already have a well-specified Bayesian model form for the problem at hand; however, this may be challenging in arbitrarily large and complex hypothesis spaces.
In particular, their specific method is not applicable more widely beyond the preference learning context they consider.

\paragraph{BED}
It has been noted that the traditional sequential BED approach can sometimes be suboptimal in practice, as it only optimizes the EIG of the next observation, without planning ahead for the fact that design decisions taken at a given step can also influence the achievable EIGs from future steps~\citep{foster2021variational}.
A variety of \emph{policy--based} BED approaches have subsequently been proposed to address this~\citep{foster2021deep,ivanova2021implicit,blau2022optimizing,huan2016sequential,hedman2025stepdad}, while also removing the need to make model updates and conduct optimizations during the experiment itself.
Our findings are complementary: by providing more faithful model factorizations, belief updates, and EIG estimators in the LLM setting, \name could supply stronger building blocks for policy-based methods, reducing variance, enhancing effectiveness, and improving the sample efficiency of policy training.

\begin{algorithm}[tb]
\caption{Data--Estimation sequential information gathering}
\label{alg:sequential-eig}
\begin{algorithmic}[1]
\Require LLM $\pmodel$, history $h_0 = \varnothing$, budget $T$, num.\ hypothesis samples $M$
\For{$t = 1, \ldots, T$}
    \State \textbf{Generate candidates:} sample $\mathcal{X}_{\mathrm{cand}} = \{x_t^{(1)}, \ldots, x_t^{(K)}\}$ from $\pmodel(x_t; h_{t-1})$
    \For{each candidate $x_t \in \mathcal{X}_{\mathrm{cand}}$}
        \For{each $y \in \mathcal{Y}$} \algcomment{enumerate answer options}
            \State $w_y \leftarrow \pmodel(y; [h_{t-1}, x_t])$ \algcomment{answer prob.\ via LLM logits}
            \State Sample $\{\theta_m^{(y)}\}_{m=1}^{M} \sim \pmodel(\theta; [h_{t-1}, x_t, y])$
            \State Estimate $\hat{H}_y \leftarrow \widehat{\mathrm{H}}\!\left[\pmodel(\theta; [h_{t-1}, x_t, y])\right]$ using $\{\theta_m^{(y)}\}$
        \EndFor
        \State $\hat{H}_{\mathrm{cond}} \leftarrow \sum_{y \in \mathcal{Y}} w_y \, \hat{H}_y$
        \algcomment{Term 1: marginal entropy of $\theta$ (design-dependent)}
        \State \label{line:de-marginal-start} Pool: $\{\theta_m\}_{m=1}^{M'} \leftarrow \bigcup_{y \in \mathcal{Y}} \{\theta_m^{(y)}\}$ \algcomment{samples from mixture}
        \For{$m = 1, \ldots, M'$} \label{line:de-cost}
            \State $\ell_m \leftarrow \log \sum_{y \in \mathcal{Y}} w_y \cdot \pmodel(\theta_m; [h_{t-1}, x_t, y])$
        \EndFor
        \State \label{line:de-marginal-end} $\hat{H}_{\mathrm{marg}} \leftarrow -\frac{1}{M'}\sum_{m=1}^{M'} \ell_m$
        \State $\widehat{\mathrm{EIG}}(x_t) \leftarrow \hat{H}_{\mathrm{marg}} - \hat{H}_{\mathrm{cond}}$
    \EndFor
    \State \textbf{Select:} $x_t^* \leftarrow \arg\max_{x_t \in \mathcal{X}_{\mathrm{cand}}} \widehat{\mathrm{EIG}}(x_t)$
    \State \textbf{Ask and observe:} pose $x_t^*$, receive $y_t$, set $h_t \leftarrow (h_{t-1}, (x_t^*, y_t))$
\EndFor
\end{algorithmic}
\end{algorithm}

\section{Data--estimation method}
\label{sec:data_estimation_method}

Our \emph{Data--Estimation} method is based on a model derived from a data--estimation pairing (\Cref{sec:method}).
\begin{align}
    \label{eq:de-joint} p(\theta, \data_t; \history, \design_t) = \yllm \postllm.
\end{align}

\subsection{EIG estimation}

Using the form of the EIG given by \Cref{eq:bayes_eig_theta}, we have
\begin{align}
    \label{eq:de-eig-full}
    \eig_{\theta}(\design) 
    = \underbrace{\mathrm{H}\!\left[p(\theta; \history, \design)\right]}_{\text{marginal entropy of } \theta}
    \;-\; \underbrace{\mathbb{E}_{\yllm}\!\left[\mathrm{H} \left[\postllm\right]\right]}_{\text{expected posterior entropy}}.
\end{align}
Crucially, the marginal on $\theta$ implied by this pairing is obtained by integrating out $\data$:
\begin{align}
    \label{eq:de-marginal}
    p(\theta; \history, \design) = \sum_{\data_t} \yllm \postllm,
\end{align}
which is a mixture distribution whose components and weights both depend on $\design$.
Unlike in the prior--likelihood pairing, where the belief state $p_f(\theta; \history)$ is constructed independently of the candidate design, the marginal entropy $\mathrm{H}[p(\theta; \history, \design)]$ here \emph{varies across candidate questions and cannot be dropped} from the optimization.
This is a concrete manifestation of the issue identified in \Cref{sec:method}: the data--estimation pairing does not provide a design--independent belief state on $\theta$.

\paragraph{Discrete answer space.}
When the possible values for $\data$ are discrete (and finite) and we can evaluate $\yllm$ in closed form, both terms of \Cref{eq:de-eig-full} can be computed directly.
The second term (expected posterior entropy) is:
\begin{align}
    \label{eq:de-cond-entropy}
    \mathbb{E}_{\yllm}\!\left[\mathrm{H}\!\left[\postllm\right]\right]
    = \sum_{\data_t} \yllm \mathrm{H}\!\left[\postllm\right],
\end{align}
where each posterior entropy $\mathrm{H}[\postllm]$ can be evaluated from the LLM's logits or, if these are unavailable, estimated by sampling.
The first term (marginal entropy) requires evaluating $p(\theta; \history, \design)$ as the mixture in \Cref{eq:de-marginal}, then computing its entropy.
In practice, this entropy cannot be obtained in closed form and must be estimated.
A natural approach is to draw samples $\theta_m \sim p(\theta; \history, \design)$ (by first sampling $\data_m \sim \yllm$, then $\theta_m \sim \postllm[,\data_m]$) and using a plug-in or nearest-neighbor entropy estimator, or evaluating $\log p(\theta_m; \history, \design) = \log \sum_{\data} \yllm\, \pmodel(\theta_m; [\history, \design, \data])$ exactly by enumerating over $\data$.

\paragraph{Monte Carlo estimator.}
When $\data$ cannot be enumerated, both terms must be estimated by Monte Carlo.
Drawing $N$ samples $\data_n \sim \yllm$, the full estimator takes the form:
\begin{align}
    \label{eq:de-mc-full}
    \eig_\theta(\design; \history) \approx \;
    \mathrm{H}\!\left[\frac{1}{N}\sum_{n=1}^{N} \pmodel(\theta; [\history, \design, \data_n])\right]
    \;-\; \frac{1}{N}\sum_{n=1}^{N} \mathrm{H}\!\left[\pmodel(\theta; [\history, \design, \data_n])\right].
\end{align}
Whereas \Cref{eq:est-rb} performs this computation over the answer space $\mathcal{Y}$ (which is small and enumerable for multiple-choice questions), \Cref{eq:de-mc-full} requires it over the hypothesis space $\Theta$.
For the first term, evaluating the mixture entropy requires computing $\pmodel(\theta; [\history, \design, \data_n])$ for a set of sampled $\theta$ values across all $N$ mixture components --- an $O(M \times N)$ cost in autoregressive probability evaluations over the full hypothesis text.

\subsection{Implementation}
\label{sec:de-implementation}

We provide an overview of the Data--Estimation procedure in \Cref{alg:sequential-eig}.
In our experiments, the answer space $\mathcal{Y}$ is enumerable (multiple-choice options), so we use \Cref{eq:de-cond-entropy} for the second term.
For the first term, we estimate the marginal entropy by sampling $\theta$ values from the mixture in \Cref{eq:de-marginal} and evaluating $\log p(\theta; \history, \design)$ by enumerating over $\data$; that is, we use:
\begin{align}
    \label{eq:de-marginal-entropy-est}
    \mathrm{H}\!\left[p(\theta; \history, \design)\right] 
    \approx -\frac{1}{M} \sum_{m=1}^{M} \log \!\left(\sum_{\data} \yllm \, \pmodel(\theta_m; [\history, \design, \data])\right),
\end{align}
where $\theta_m$ are drawn from $p(\theta; \history, \design)$.
This requires $M \times |\mathcal{Y}|$ evaluations of $\pmodel(\theta_m; [\history, \design, \data])$ per candidate question, each involving a full forward pass over the hypothesis text.

\subsection{Generating candidate hypotheses}
To generate candidate values of $\theta$ for the data--estimation method, we use the prompt in \Cref{fig:theta_prompt}. 

\Cref{fig:theta_dist} shows an example of the distribution of samples obtained following two rounds in the 20--questions game. 
Note that the samples are highly concentrated on just a handful of answers. This lack of diversity shows that the model's belief distribution is far more concentrated relative to the variability over valid hypotheses in the ground--truth task distribution, 
which negatively impacts the performance of the data--estimation method.

\begin{figure}[tb] 
    \centering
    \begin{tcolorbox}[
        colback=blue!4, colframe=blue!40!black, 
        title=Conversation History ($\mathcal{D}$), 
        fonttitle=\bfseries\scriptsize, fontupper=\scriptsize,
        left=4pt, right=4pt, top=3pt, bottom=3pt, boxsep=1pt,
        toptitle=1.5pt, bottomtitle=1.5pt
    ]
    \textbf{Q:} Is this person known for their contributions to science? \hspace{2pt} \textbf{A:} No.\\
    \textbf{Q:} Is this person known for their contributions to the arts? \hspace{8pt} \textbf{A:} Yes.
    \end{tcolorbox}
    
    \vspace{-0.6em} %
    
    \begin{tcolorbox}[
        colback=gray!5, colframe=gray!60!black, 
        title=Sampled Hypothesis Distribution, 
        fonttitle=\bfseries\scriptsize, fontupper=\scriptsize,
        left=4pt, right=4pt, top=4pt, bottom=4pt, boxsep=1pt,
        toptitle=1.5pt, bottomtitle=1.5pt
    ]
    \centering
    \begin{tabular}{@{}r@{\hspace{2ex}}r@{\hspace{1ex}}l@{}}
    Vincent van Gogh & 93 & \textcolor{gray!70}{\rule{3.00cm}{0.7em}} \\
    Salvador Dali    & 44 & \textcolor{gray!70}{\rule{1.42cm}{0.7em}} \\
    Frida Kahlo      & 37 & \textcolor{gray!70}{\rule{1.19cm}{0.7em}} \\
    Georgia O'Keeffe & 10 & \textcolor{gray!70}{\rule{0.32cm}{0.7em}} \\
    August Wilson    &  8 & \textcolor{gray!70}{\rule{0.26cm}{0.7em}} \\
    Auguste Rodin    &  8 & \textcolor{gray!70}{\rule{0.26cm}{0.7em}} \\
    \end{tabular}
    \end{tcolorbox}
    
    \caption{
        An empirical hypothesis distribution generated by the LLM, conditioned on the conversation history (top). By independently sampling $N=200$ hypotheses, we observe significant \textit{mode collapse}; the probability mass is heavily concentrated on a few prominent figures (e.g., Vincent van Gogh). This overconfidence negatively impacts the performance of the data-estimation baseline. Note that this aggregated view is purely diagnostic; \Cref{alg:sequential-eig} operates directly on the probabilities of individual samples.
    }
    \label{fig:theta_dist}
\end{figure}

\begin{promptbox}{Prompt for generating hypotheses (and evaluating their probability) for the data--estimation method.\label{fig:theta_prompt}} 
Return only the full name of one randomly selected famous person (living or deceased) consistent with the questions and answers above. \\

To increase randomness:\\
1. Internally brainstorm a pool of diverse and representative individuals. \\
2. Avoid defaulting to the most globally ubiquitous celebrities or famous figures. \\

Output rules: \\
- Output ONLY the person's full name (with spaces, capitalization and accents), nothing else. \\
- No extra words, explanations, numbering, or punctuation beyond what's in the name itself (hyphens/apostrophes allowed if part of the name).
\end{promptbox}
 \vspace{10pt}

\Cref{fig:icl_fails} shows an example of in-context belief updating failing to respect the conversation history. The response to the first question indicates that the target is male; however, after the second question, the LLM generates two female candidate hypotheses.
 \vspace{10pt}

\begin{promptbox}{An example of in-context belief tracking with GPT-4o-mini proposing hypotheses inconsistent with the history.\label{fig:icl_fails}} 
Q1:  Is the person male? \\
Answerer:  Yes. \\

Q2:  Is this person often associated with civil rights or social justice? \\
Answerer:  Yes. \\

Sampled hypotheses:  ['James Baldwin', 'A. Philip Randolph', 'Angela Davis', 'Malcolm X', 'W.E.B. Du Bois', 'Desmond Tutu', 'Cesar Chavez', 'Rosa Parks', 'Martin Luther King Jr.', 'Langston Hughes', 'John Lewis', 'Frederick Douglass', 'Nelson Mandela', 'Thurgood Marshall', 'Bayard Rustin']\\

Hypotheses rejected with filtering: ['Angela Davis', 'Rosa Parks']
\end{promptbox}

\section{Generating candidate hypotheses for \name}
\label{sec:candidate_hypotheses}

\begin{promptbox}{Prompt for generating candidate hypotheses for the ``Things'' dataset. Similar prompts were used for ``Celebrities'' and ``Animals''.\label{fig:things_prompt}} 
You are playing a game of 20 Questions. Using all of the questions and answers so far: \\   

Generate up to \{num\_samples\} candidate entities that satisfy every clue. \\
Each candidate must be a single, self-contained entity (e.g., "Europa", "Bagpipe", "Diadem"). \\
List each entity on its own line - no numbering, punctuation, or extra text. \\
Produce a varied set by identifying features not implied by the clues and diversifying along them. \\
Do not repeat any entity. \\

Return only the list of entities.
\end{promptbox}

A fundamental challenge for \name and its ablations is generating a sufficiently diverse set of candidate hypotheses from the LLM's belief distribution, that are consistent with the previously-answered questions. Below, we detail the steps we take to construct our distribution over hypotheses.

\paragraph{Candidate hypotheses are generated jointly, rather than independently} As illustrated in~\Cref{fig:theta_dist}, the raw distribution $\pllm(\theta)$ is highly overconfident, often concentrating mass on only a few high--likelihood hypotheses. Thus, it is not practical to directly use the LLM's distribution as a prior $p(\theta)$.
Instead, we jointly sample candidates $\theta$ and assume a uniform distribution over them.
We can view this as sampling $\theta$s from a mixture distribution. The LLM is prompted to generate a list of $N$ hypotheses in a single rollout, which corresponds to drawing from the autoregressive list distribution
\[
\pllm(\theta^{(1)}_t, \ldots, \theta^{(N)}_t ; h_t) \;=\; \prod_{n=1}^{N} \pllm(\theta_t^{(n)}; [\theta_t^{(1:n-1)}, h_t])\,.
\]

\paragraph{We use a diversity-encouraging prompt} We use a prompt designed to elicit stratified hypotheses by encouraging the LLM to consider different semantic features (e.g.\ age groups, genres, or categories) and implicitly diversify across them. An example prompt is shown in \Cref{fig:things_prompt}. In our generation prompt, we reverse the order of the question--answer pairs in $h_t$ to place the most recent question at the top of the context window (while retaining earlier exchanges), ensuring that specific constraints are prioritized and mitigating context drift. For the 20 Questions experiments, we used a higher-than-normal temperature ($T=1.3$) to increase diversity of responses. For the preference elicitation experiments, we used $T=1$ to obtain more coherent responses.

\paragraph{Candidates are filtered based on the history} For each candidate, we use $\pllm(\theta;\history)$ to assess whether it is compatible with the previous question/answer pairs. We filter responses where the probability of the given answer falls below a certain threshold; in our experiments we set this threshold to 0.2.

\paragraph{Valid candidates from previous generations are included} We also filter the candidate hypotheses from the previous generation, based on the most recent question/answer pair, and include these in our candidate set.
 We repeat the generation process, keeping the previously generated and filtered samples in context to elicit new generations, either twice or three times if sufficient hypotheses have not been generated (noting the number of possible valid samples can be less than the number requested).

\paragraph{We define a uniform prior over hypotheses}
While one could in principle reweight candidates using importance sampling, in practice we choose to not rely on the model's internal probabilities. Instead we define the prior as a uniform distribution:
\[
p_f(\theta; h_t) \approx \frac{1}{|\Theta|} \sum_{\theta \in \Theta} \delta_\theta.
\]

Finally, we note that different LLMs respond differently to strategies aiming to increase diversity: some benefit more from a higher temperature while others benefit from more repetitions of the sampling--filtering cycle. For fairness, in our experiments we have kept these parameters constant across models. %

\section{Additional experimental results}
\label{app:experimental_results}

\subsection{20 Questions and Preference Elicitation}
\begin{figure}[h]
    \includegraphics[width=\linewidth]{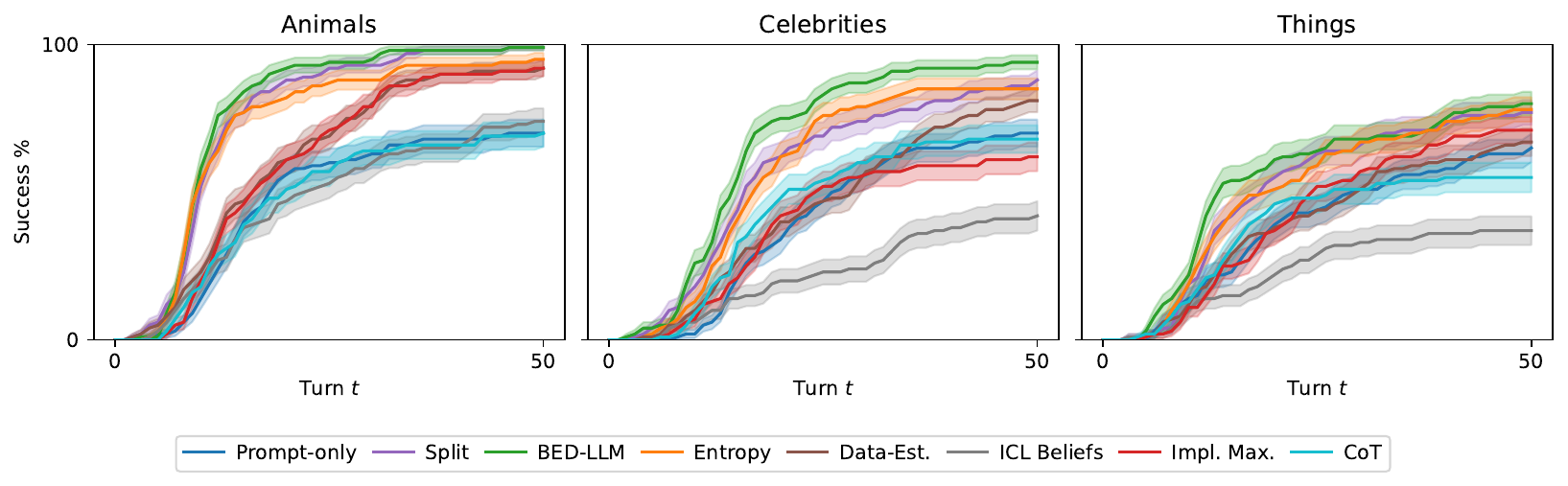}
    \caption{
        Success rate on 20 Questions with Qwen2.5-72B, extended to 50 turns: mean $\pm$ standard error across 100 targets per dataset. BED-LLM remains the best-performing method at every turn. The performance gap to baselines narrows in later turns as the task saturates --- most methods eventually solve the easier instances --- but BED-LLM consistently retains its advantage.
    }
    \label{fig:20q_50turns}
\end{figure}

\begin{figure}[h]
    \centering
    \includegraphics[width=0.6\linewidth]{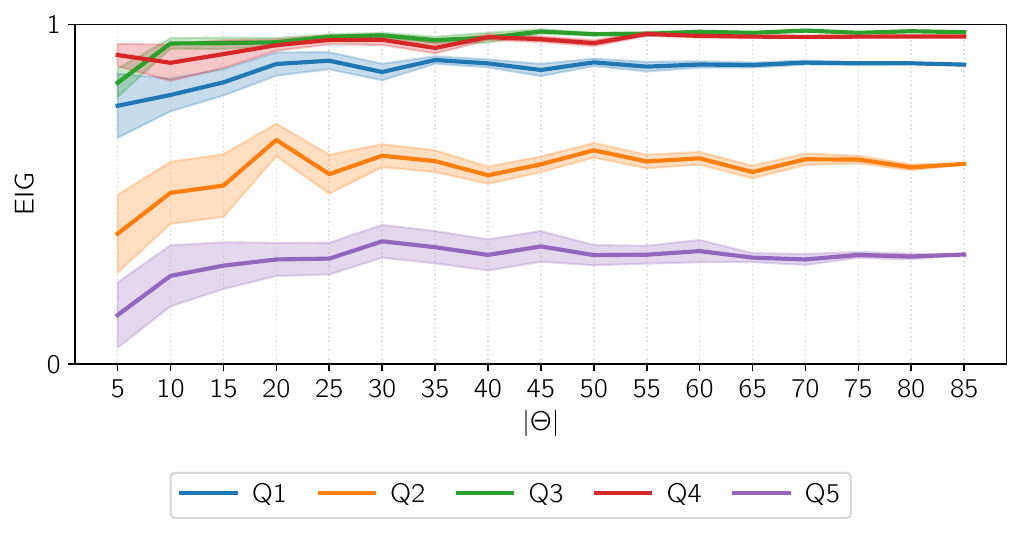}
    \caption{
        Analysis of the empirical convergence of the Rao-Blackwellized EIG estimator using GPT-4o-mini.
        We plot EIG estimates for 5 candidate questions at turn 5 of the 20 Questions game with $\theta^*=$``Saoirse Ronan'', as a function of the number of samples in the hypothesis set $\Theta$.
    }
    \label{fig:eig_questions}
\end{figure}

\Cref{fig:eig_questions} shows the EIG estimates for five candidate questions
as the number of hypothesis samples $|\Theta|$ increases, at turn~5 of a
representative 20 Questions game. The estimates converge rapidly with the number of samples, and more importantly, the \emph{ranking} of candidate
questions stabilizes well before the estimates themselves have fully converged.
By $|\Theta| = 10$ the relative ordering of all five candidates already matches
the ordering observed at $|\Theta| = 85$. This means that the question
selection decision---which is what matters algorithmically, since we only
need the $\arg\max$, not the absolute EIG value---is robust to using a small
hypothesis set. This justifies \name's use of a moderate
sample budget without sacrificing decision quality.

\begin{figure}[tb]
    \includegraphics[width=\linewidth]{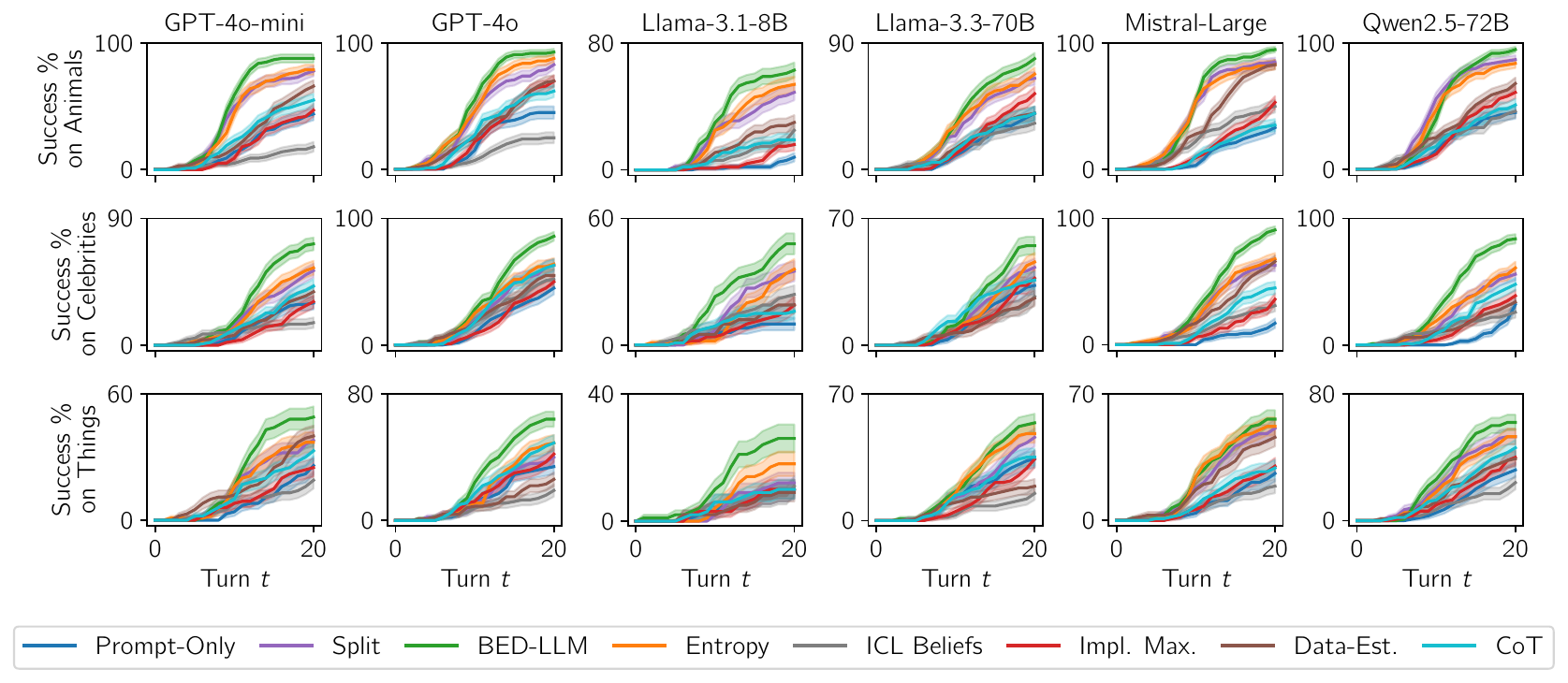}
    \caption{
        Success rate on 20 Questions.
        Mean $\pm$ standard error across 100 targets. \name beats all other methods across all datasets and models evaluated.
    }
    \label{fig:full_react}
\end{figure}

\begin{figure}[tb]
    \centering
    \includegraphics[width=\linewidth]{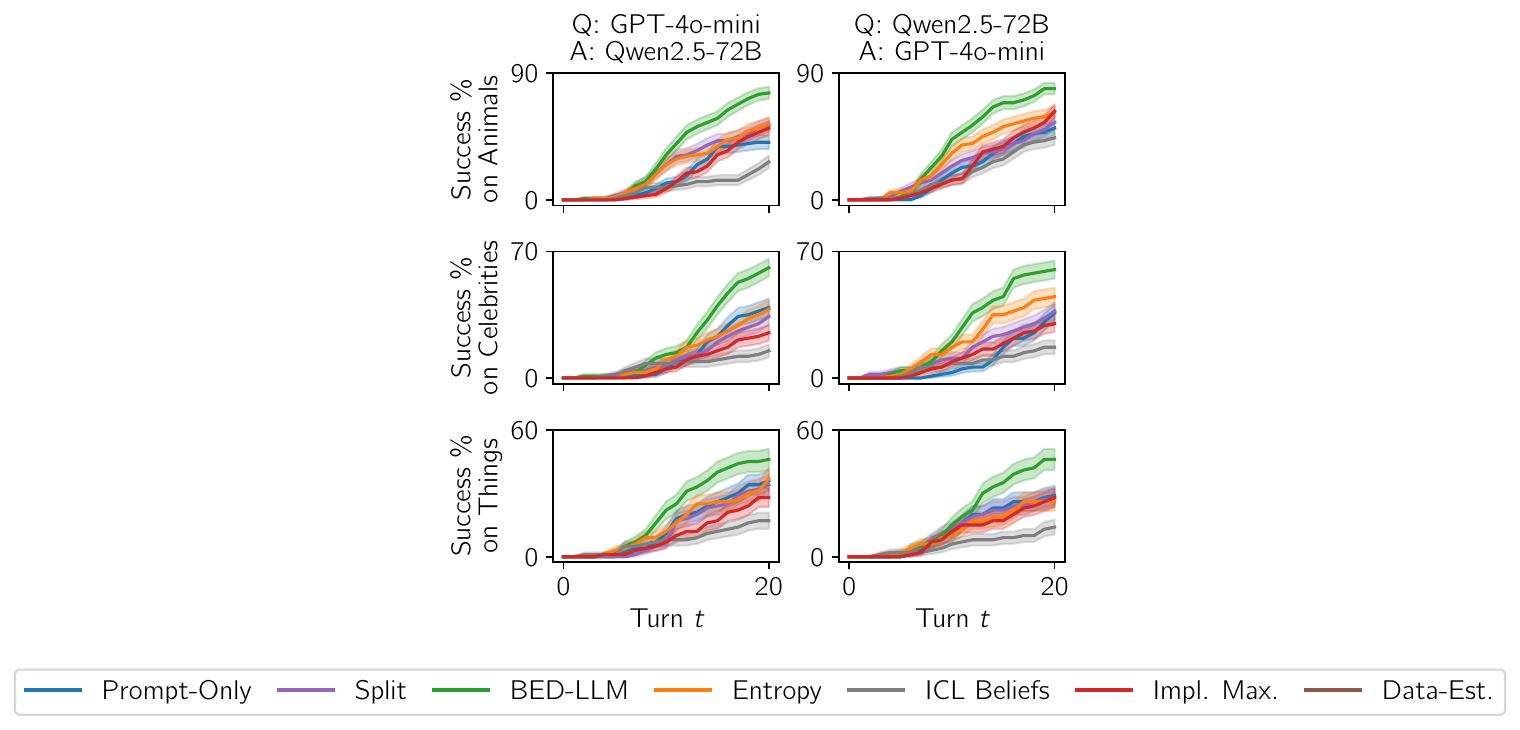}
    \caption{Full plots for 20 Questions Experiments with different questioner and answerer.}
    \label{fig:different_models}
\end{figure}

\begin{figure}[tb]
    \includegraphics[width=\linewidth]{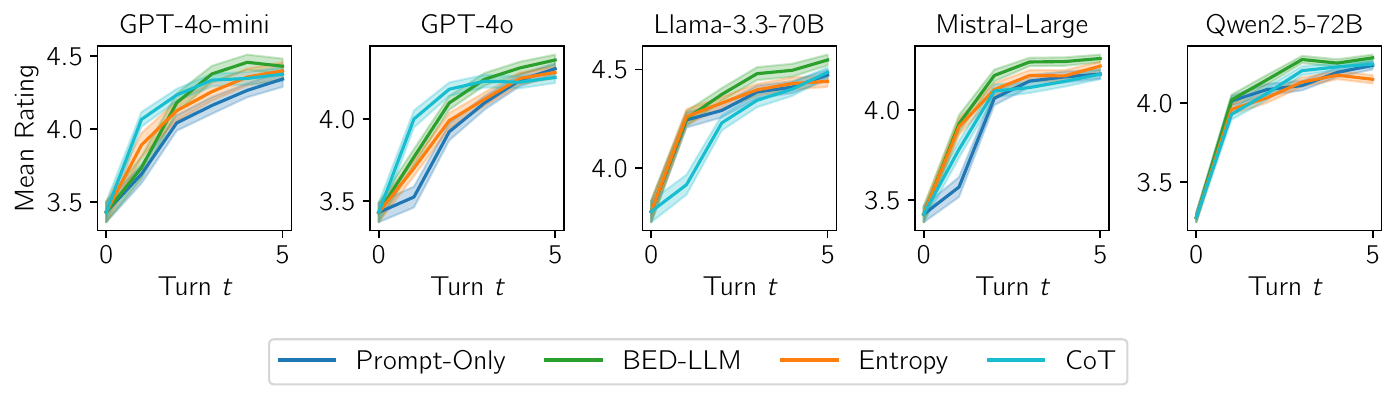}
    \caption{
        Mean rating across 10 film recommendations: mean $\pm$ standard error across 200 users. \name beats all other methods across all datasets and models evaluated.
    }
    \label{fig:full_preferences}
\end{figure}

\subsection{Wall-clock times}
\begin{table}[h]
\centering
\begin{tabular}{lr}
\hline
Method & Runtime \\
\hline
Data--Estimation & 1d 17h 13m 34s \\
ICL Beliefs & 3h 10m 39s \\
Entropy & 2h 46m 19s \\
Split & 2h 30m 35s \\
BED-LLM & 2h 28m 43s \\
Implicit Maximization & 45m 10s \\
CoT & 35m 28s \\
\baseline & 26m 29s \\
\hline
\end{tabular}
\caption{Wall-clock runtimes 
for 20 Questions for Qwen2.5-72B on the entire \emph{Animals} problem set.}
\label{fig:wall-clock}
\end{table}

\Cref{fig:wall-clock} shows that Data--Estimation is over an order of magnitude slower than all other methods: 
its full-sequence likelihood evaluations per candidate question dominate the cost.
Among the remaining methods, \name's runtime is comparable to---and even slightly lower than---Entropy and Split, 
despite computing the full EIG; 
this is because the additional conditional-entropy term in \Cref{eq:est-rb} reuses the same likelihood evaluations 
and adds negligible overhead.
\baseline and CoT are the cheapest methods, as they require only a single LLM generation call per turn 
with no hypothesis sampling or objective computation.
Implicit Maximization falls in between: it samples hypotheses and candidate questions 
but replaces explicit EIG estimation with a single LLM selection call.
ICL Beliefs is slightly more expensive than BED-LLM because the absence of filtering 
leads to lower-quality hypotheses, requiring a much larger number of questions before the game terminates.

\subsection{Effect of sequentially updating the \name likelihood}
\label{app:updating_likelihood}
As discussed in~\cref{app:updating_likelihood}, we choose to use a \emph{static} likelihood, $\seqlikellm$, in the model factorization for \name.
Alternatively, we could condition the likelihood on the full interaction history, i.e.~$\pllm(\data_{t+1};[\history,\theta,\design_{t+1}])$. However, this can lead to undesirable effects (e.g.~context-induced calibration shifts). We provide results in~\cref{tab:history_summary} for the 20 Questions game where we update the likelihood, and demonstrate that it leads to consistently worse performance than \name, supporting our rationale in~\cref{app:updating_likelihood} to keep the likelihood static.

\label{app:name_summary_history}
\begin{table}[h]
  \centering
  \begingroup
  \setlength{\tabcolsep}{6pt}
  \renewcommand{\arraystretch}{1.05}
  \footnotesize
  \begin{tabular}{llrr}
    \toprule
    \multicolumn{2}{c}{} & \multicolumn{2}{c}{\textbf{Success Rate (\%)}} \\
    \cmidrule(l){3-4}
    \textbf{Dataset} & \textbf{Model} & {\textbf{\name + static likelihood}} & {\textbf{\name + likelihood updating}} \\
    \midrule
    \multirow{5}{*}{Animals}
      & GPT-4o-mini    & 88 $\pm$ 3.3 & 85 $\pm$ 3.6 \\
      & GPT-4o         & 93 $\pm$ 2.6 & 86 $\pm$ 3.5 \\
      & Llama-3.1-8B   & 63 $\pm$ 4.9 & 67 $\pm$ 4.7 \\
      & Llama-3.3-70B  & 79 $\pm$ 4.1 & 80 $\pm$ 4.0 \\
      & Qwen2.5-72B    & 95 $\pm$ 2.2 & 89 $\pm$ 3.1 \\
    \midrule
    \multirow{5}{*}{Celebrities}
      & GPT-4o-mini    & 72 $\pm$ 4.5 & 63 $\pm$ 4.9 \\
      & GPT-4o         & 86 $\pm$ 3.5 & 83 $\pm$ 3.8 \\
      & Llama-3.1-8B   & 58 $\pm$ 5.0 & 56 $\pm$ 5.0 \\
      & Llama-3.3-70B  & 55 $\pm$ 5.0 & 44 $\pm$ 5.0 \\
      & Qwen2.5-72B    & 84 $\pm$ 3.7 & 70 $\pm$ 4.6 \\
    \midrule
    \multirow{5}{*}{Things}
      & GPT-4o-mini    & 49 $\pm$ 5.0 & 43 $\pm$ 5.0 \\
      & GPT-4o         & 64 $\pm$ 4.8 & 57 $\pm$ 5.0 \\
      & Llama-3.1-8B   & 26 $\pm$ 4.4 & 24 $\pm$ 4.3 \\
      & Llama-3.3-70B  & 55 $\pm$ 5.0 & 54 $\pm$ 5.0 \\
      & Qwen2.5-72B    & 62 $\pm$ 4.9 & 61 $\pm$ 4.9 \\
    \bottomrule
  \end{tabular}%
  \endgroup
  \caption{\name success rates across datasets and models. $\pm$ numbers show the standard error of the mean 
    estimated using $\sqrt{p(1-p)/(n-1)}$ where $p$ is the success percentage and $n$ is the number of datapoints. 
    This estimator is positively biased and thus conservative.}
  \label{tab:history_summary}
\end{table}

\FloatBarrier

\section{Further description of methods and ablations}
\label{app:ablations}

\textbf{\baseline}:
A direct prompting baseline where the model is asked to immediately generate the next question given the dialogue history, no candidate generation, and no belief modeling (\Cref{fig:naive_prompt}). This corresponds to the simplest and most common way users currently interact with LLM agents for question-asking tasks, and therefore serves as a natural “standard LLM prompting’’ baseline. After the LLM has generated its question, it waits for the user response and generates a new question based on the new history.

\begin{promptbox}{Prompt for the \baseline method.\label{fig:naive_prompt}}
Your task is to ask a series of questions to deduce the identity of the {'person' if task == 'celebrities' else 'animal' if task == 'animals' else 'entity'} that I'm thinking of with as few queries as possible.\\
Only ask questions that can be answered by 'yes', 'no' or 'maybe'.\\
Do not ask for hint. Make your question standalone with no linebreaker.\\

Output format: <question>Your question here?</question>\\

Now start asking a question.
\end{promptbox}

\textbf{CoT}:
A direct prompting baseline where the model is asked to reason and provide action (\Cref{fig:cot_prompt}) loosely following the ReAct framework \citep{yao2023react}.

\begin{promptbox}{Prompt for the CoT method.\label{fig:cot_prompt}}
Your task is to ask a series of questions to deduce the identity of the {'person' if task == 'celebrities' else 'animal' if task == 'animals' else 'entity'} that I'm thinking of with as few queries as possible. \\
Only ask questions that can be answered by 'yes', 'no' or 'maybe'.\\
Do not ask for hints. The actual question you ask must be standalone with no linebreakers.\\

Use the following format for every response:\\
Thought: <brief reasoning about what to ask next>\\
Action: ASK["<a single yes/no/maybe question with no linebreaks>"]

Now start by producing your first Thought and Action.
\end{promptbox}

\textbf{Split}:
Essentially a version of our method in which we cast previous approaches, such as \cite{kobalczyk2025active}:
uses the same candidate question generating method as BED-LLM,
assumes a deterministic likelihood over hypotheses (i.e., each hypothesis deterministically predicts a single answer to each question), and
Evaluates the question on the sampled hypotheses, and selects the question that maximally balances the hypotheses, i.e., splits the current hypothesis set into subsets whose sizes are as close to equal as possible.

\paragraph{Implicit Maximization (IM).}
The \textit{Implicit Maximization} baseline is a lightweight, reasoning-driven method inspired by the \textit{Tree-of-Thoughts} \citep[ToT,][]{yao2023tree} framework. ToT methods explicitly expand a search tree of intermediate reasoning steps (``thoughts'') before selecting an action. Implicit Maximization provides a computationally efficient, collapsed variant of this idea:
\begin{itemize}
    \item Rather than explicitly branching over thoughts, the model is prompted to internally deliberate over several possible next questions and their consequences.
    \item The branching and evaluation occur inside the model’s chain-of-thought, rather than through externally enumerated tree expansion.
    \item The model then outputs the question it judges to be most informative, performing an amortized, ToT-style lookahead within a single LLM call.
\end{itemize}

Thus, IM captures the central intuition of Tree-of-Thoughts—reasoning over hypothetical futures—while avoiding the substantial computational cost of explicit tree search. It serves as a strong inference-time reasoning comparator to BED-LLM.

\section{Experiment details for 20 Questions}\label{app:20q_details}

\subsection{Problem sets}\label{app:20q_datasets}

We evaluate across three distinct problem sets---Animals, Celebrities, or Things---with each containing a mix of 100 obscure and common targets.
Here, the problem set is just a list of different $\theta^*$ that will be individually provided to the answerer to instantiate different problems (e.g.~we conduct a trial where $\theta^*=$``dog'', then one where $\theta^*=$``cat'', etc).
The list of targets is \emph{never} provided to the questioner model to restrict the set of possible hypotheses:
the questioner is only prompted that is trying to identify an ``animal'', ``celebrity'', or ``thing''.
The problem sets are

\begin{itemize}
    \item Animals: a set of animal species generated with OpenAI's o3 model to ensure a diverse mix and balanced taxonomy.
    \item Celebrities: a diverse set of public figures, as used by \citet{zhang2024probingmultiturnplanningcapabilities}.
    \item Things: a collection of everyday and exotic entities drawn from the web corpus, as used by \citet{zhang2024probingmultiturnplanningcapabilities}; it covers a wide range of categories, from plants and clothing to professions, events, and mythical creatures.
\end{itemize}

To create the Animals problem set, we prompted OpenAI o3~\citep[\texttt{o3-2025-04-16}]{openai_o3_2025} to generate a list of animals, using the prompt in \Cref{fig:animals_prompt}.
The resulting list is shown in \Cref{fig:animal_dataset}. 
Alternative names (after $|$ character) were manually added.

\vspace{3mm}

\begin{promptbox}{Prompt for Animals problem set generation.\label{fig:animals_prompt}} 
You are a zoologist.\\
Please generate a list of 100 living animal species with very high taxonomic diversity, including diversity in phyla, classes, orders, and families. Present each animal on a different line. 
\end{promptbox}

\begin{figure}[h]
    \begin{multicols}{3}
        African elephant\\
        Bengal tiger\\
        Bald eagle\\
        Blue whale\\
        Red kangaroo\\
        Giant panda\\
        Snow leopard\\
        Green sea turtle\\
        American alligator\\
        Bottlenose dolphin\\
        Emperor penguin\\
        Great white shark\\
        Golden poison frog | Golden poison dart frog\\
        Honey bee\\
        Monarch butterfly\\
        Okapi\\
        Chimpanzee\\
        Arctic fox\\
        Komodo dragon\\
        Giraffe\\
        Cheetah\\
        Hammerhead shark\\
        Axolotl\\
        Orca\\
        Puffin\\
        Red panda\\
        Platypus\\
        Rhinoceros beetle\\
        Tasmanian devil\\
        Wombat\\
        Sloth\\
        Blue-ringed octopus | Blue ringed octopus\\
        Manatee\\
        Narwhal\\
        Sea otter\\
        Coral snake\\
        King cobra\\
        Harpy eagle\\
        Lemur\\
        Koala\\
        Aye-aye | Ayeaye\\
        Snowy owl\\
        Elk\\
        Wolverine\\
        Caracal\\
        Cassowary\\
        Quokka\\
        Pangolin\\
        Saiga antelope\\
        Galápagos tortoise | Galapagos tortoise\\
        Sumatran orangutan\\
        Red-eyed tree frog | Redeyed tree frog\\
        European badger\\
        Moose\\
        African grey parrot\\
        Scarlet macaw\\
        Black mamba\\
        Albatross\\
        Humpback whale\\
        Dugong\\
        Anaconda\\
        Kookaburra\\
        Coyote\\
        Brown bear\\
        Golden jackal\\
        Capybara\\
        Ibex\\
        Japanese macaque\\
        Kiwi\\
        Leafcutter ant\\
        Mantis shrimp\\
        Ocelot\\
        Peregrine falcon\\
        Quetzal\\
        Raccoon\\
        Sand cat\\
        Tarantula\\
        Uakari\\
        Vicuña\\
        Wildebeest\\
        Rock hyrax | dassie\\
        Yak\\
        Zebra\\
        Blue dragon nudibranch | Blue dragon sea slug\\
        Chinchilla\\
        Dhole\\
        Electric eel\\
        Flying fox\\
        Gharial\\
        Horseshoe crab\\
        Indigo bunting\\
        Jerboa\\
        Kakapo\\
        Lionfish\\
        Markhor\\
        Nautilus\\
        Olive baboon\\
        Pika\\
        Quoll\\
        Rosy boa\\
    \end{multicols}
\caption{Animals problem set (generated using OpenAI o3, with manual curation)}
\label{fig:animal_dataset}
\end{figure}

\subsection{Evaluation}

We assess performance by tracking the questioner's ability to identify the hidden target $\theta^*$ over the course of each game.
At each turn $t$, we prompt it to produce a single guess for $\theta^*$ via greedy decoding---that is, we extract the highest likelihood candidate from the belief state of the questioner $p_f(\theta; h_t)$. 
This guess is evaluated against the true target $\theta^*$ (including alternative names) using case--insensitive exact string matching and we measure the proportion of correct guesses at each turn. 
Importantly, these evaluation guesses are \emph{not} part of the questioner algorithm itself: they are extractions from the questioner's belief state $p_f(\theta; h_t)$ and are excluded from $\history$ to not affect subsequent question selection.  
In line with the original rules of the game, we also introduce an explicit mechanism for the questioner to guess the answer as part of its 20 questions: if the set of filtered hypotheses collapses to a single candidate, or a direct guess of $\theta^*$ is evaluated as the maximally informative question by the acquisition function, the questioner asks ``Is it $\langle$\texttt{item}$\rangle$?''. This guess is evaluated using exact string matching, as above. If there is a match, the game terminates successfully; otherwise, if $t<20$, the game continues with the question and negative response included in $\history$ and counted towards the 20 question budget.

\subsection{Algorithmic details}

Using our sample--then--filter process (see~\Cref{sec:sBED}), we aim to sample at least $N=15$ hypotheses, repeating the cycle up to three times if needed (the exact number of hypotheses can be less than this as it may not be possible to generate sufficient valid hypotheses, especially in later experiment turns).
The questioner generates $M=15$ candidate questions to test, $\mathcal{X}^{\text{cand}}$, using the ``conditional generation'' approach of~\Cref{sec:q-gen} when possible, but falling back on ``unconditional generation'' if insufficient candidate hypotheses have been generated.

\section{Experiment details for preference elicitation}\label{app:preference_details}

\subsection{Problems}

To generate a set of ground-truth user profiles, we take a set of $200$ real user ratings from the MovieLens-100K dataset~\citep{harper2015movielens}, then use an ``oracle'' LLM (namely, OpenAI's o3 model) to produce a paragraph of text that is consistent with each distinct set of ratings, using the prompt in \Cref{fig:persona_prompt}.
As was the case for the 20 Questions problems, this problem set is never provided to the questioner and the set of allowed $\theta$ is not constrained.

Because we need the LLM to be able to meaningfully capture uncertainty in the space of responses, we restrict questions to be multiple-choice.
Specifically, the questioner is tasked with producing a question along with five possible responses A/B/C/D/E.
We then define each $x_t$ to be the question coupled with the possible responses, and each $y_t$ to be one of the letters A/B/C/D/E to provide a restricted set of tokens over which we can measure entropy.
Option E is further constrained to always be ``none of the above'' so that the answerer is not committed to choosing one of the directly generated choices if none are suitable.

\subsection{Evaluation}

To again allow tracking of progress through the experiment, after each turn of the interaction $t$, 
the questioner generates ten film recommendations, conditioned on $\history$.
These recommendations are checked for consistency with prior questions and answers; if any are judged inconsistent then they are removed and additional recommendations are generated. 
The quality of the film recommendations is then assessed using an ``LLM-as-judge'' protocol~\citep{zhu2025judgelmfinetunedlargelanguage, trivedi2024selfrationalizationimprovesllmfinegrained}. 
Namely, the answerer evaluates each of the 10 films recommended by the questioner, conditioned on the hidden target user profile $\theta^*$. 
It scores each film on a scale of 1 to 5 (in 0.5 increments), based on how well the recommendation aligns with $\theta^*$ --- this score is output together with a brief justification to increase reliability. 
We report the mean rating and standard error across 200 users, over 5 question--answer turns.

\subsection{Algorithmic details}

For \name and Entropy, we compare $M=8$ candidate questions at each turn and we aim to generate at least $N=5$ candidate hypotheses.
We use the ``unconstrained generation'' approach of candidate question generation (see~\Cref{sec:q-gen}) as the user profiles can be quite diffuse and we are only generating a small number of possible hypotheses that can be quite easy to split.

We note that data--estimation setup is not at all viable for this problem because the large number of tokens and varying dimensionality of each $\theta$ sample mean that $\mathrm{H}[\pllm(\theta;[\history,\design_{t+1},\data_{t+1}])]$ is not only infeasible to estimate, but also is not a meaningful measure of uncertainty.

\begin{promptbox}{Prompt used to generate ground-truth user profiles for preference elicitation task\label{fig:persona_prompt}.} 
You will be given a user's complete film rating history from the 
MovieLens dataset, provided as a dictionary structured by rating 
levels. \\
Your task is to thoroughly analyze the user's preferences across
the entire range of their film ratings (from highest to lowest). 
Then, write a cohesive, descriptive paragraph (approximately 5–7 
sentences) summarizing the user's overall film taste profile. \\

In your response, explicitly address:\\

Favored Elements (inferred primarily from 4.5–5.0 ratings): 

\begin{itemize}
    \item Highlight the genres, narrative styles, themes, tones, 
    historical eras, and emotional experiences that consistently 
    resonate positively 
    with this user. 
    
    \item Avoid mentioning any specific film titles, characters, or 
    explicit plot points.
    
\end{itemize}

Neutral or Mixed Preferences (inferred from ratings around 
2.5–4.0):
\begin{itemize}
    \item Note if there are indications of genre overlap or conditional 
    enjoyment, such as certain genres or styles they occasionally 
    appreciate under specific circumstances.
\end{itemize}

Disliked Elements (inferred primarily from 0.5–1.5 ratings):
\begin{itemize}
    \item Clearly outline the genres, narrative characteristics, tones, 
    or emotional impacts that the user consistently finds 
    unappealing or poorly executed.
\end{itemize}

Your paragraph must be precise, informative, nuanced, and 
balanced, effectively capturing the complexity and specificity of 
the user's movie preferences. The resulting profile should be 
clear and detailed enough for a recommendation system to 
accurately predict the user's likely enjoyment or dislike of 
other films based on their established patterns of taste. \\

Proceed carefully, reasoning explicitly about the user's overall 
rating patterns rather than relying exclusively on extreme 
ratings, to form a comprehensive, stable, and representative film 
preference profile.
\end{promptbox}

\section{Code}
Code is available at \url{https://github.com/DeeproChoudhury/BED-LLM}.

\end{document}